\documentclass[nohyperref]{article}

\usepackage{microtype}
\usepackage{graphicx}
\usepackage{subcaption}
\usepackage{booktabs} 
\usepackage{siunitx}
\usepackage[ruled,vlined, linesnumbered, noend]{algorithm2e}

\usepackage[hyperfootnotes=false]{hyperref}

\usepackage[square,numbers]{natbib}

\usepackage{neurips_data_2022}

\usepackage{amsmath}
\usepackage{amssymb}
\usepackage{mathtools}
\usepackage{amsthm}
\usepackage{listings}
\usepackage{mathrsfs}

\usepackage{amsmath,amsfonts,bm}

\def\eqref#1{equation~\ref{#1}}

\def\1{\bm{1}}

\DeclareMathAlphabet{\mathsfit}{\encodingdefault}{\sfdefault}{m}{sl}
\SetMathAlphabet{\mathsfit}{bold}{\encodingdefault}{\sfdefault}{bx}{n}

\newcommand{\cA}{\mathcal{A}}

\newcommand{\cO}{\mathcal{O}}

\newcommand{\cS}{\mathcal{S}}

\newcommand{\cX}{\mathcal{X}}

\usepackage{xcolor}
\usepackage{comment}

\renewcommand{\comment}[1]{}

\lstdefinestyle{mystyle}{
    basicstyle=\ttfamily\tiny,
}

\newcommand{\lintlong}[0]{Learning from Interaction}
\newcommand{\lint}[0]{LInt}
\newcommand{\segarlong}[0]{Sandbox Environment for Generalizable Agent Research}
\newcommand{\segar}[0]{SEGAR}

\usepackage[capitalize,noabbrev]{cleveref}

\title{The Sandbox Environment for Generalizable Agent Research (SEGAR)}

\author{%
  R Devon Hjelm\thanks{Currently at Apple Machine Learning Research}\ \ \thanks{Equal contribution}\\
  MSR \& Mila\\
  \texttt{devon.hjelm@apple.com} \\
  \And
  Bogdan Mazoure\footnotemark[2]\\ 
  McGill U. \& Mila\\
  \And
   Florian Golemo \\
   Mila \\
   \And
   Felipe V Frujeri \\
   MSR \\
   \And
   Pedro H M Braga \\
   UFPE \& ÉTS \& Mila \\
   \And
   Mihai Jalobeanu \\
   MSR \\
   \And
   Samira E Kahou \\
   ÉTS \& Mila \& CIFAR \\
   \And
   Andrey Kolobov \\
   MSR \\
}

\begin{document}
\maketitle








\vskip 0.3in



\begin{abstract}
    A broad challenge of research on generalization in sequential decision-making is designing benchmarks that clearly measure progress.
    While there has been notable headway, most popular generalization benchmarks, such as Meta-World and Procgen, consist of a fixed set of tasks. They don't provide easy access to the underlying factors of the environment that would otherwise enable a researcher to design custom source and target task distributions for evaluating particular aspects of generalization. Other benchmarks, such as CausalWorld, are more extensible but are also computationally expensive to run. 
    We built the \segarlong{} (\segar) with all of these considerations in mind. In a nutshell, \segar{} is a toolkit for defining generalization objectives over computationally lightweight sequential decision-making tasks and measuring learning algorithms' performance w.r.t. these objectives. In addition to providing several task distributions out of the box, SEGAR enables finegrained Reinforcement Learning (RL) generalization experiments by giving researchers tools for creating families of tasks relevant for testing specific hypotheses.
    We present an overview of \segar{}, explain how it contributes to generalization research, and conduct experiments illustrating a few types of research questions \segar{} can help answer.
    \segar{} is open-sourced and can be found at \href{https://github.com/microsoft/segar}{https://github.com/microsoft/segar}.
\end{abstract}

\thispagestyle{empty}
\section{Introduction}

    Consider the problem of training an automated agent to perform sequential decision-making tasks in a real-world setting, e.g.,  autonomous driving~\citep{kendall2019learning} or robotic control~\citep{levine2016end}.
Training such an agent involves learning from \emph{interaction data}, i.e., data generated as the result of an embodied actor performing actions in an environment and eliciting feedback. We refer to any setting that requires learning from data of this type, and hence entails using approaches such as Reinforcement Learning~\citep[RL,][]{sutton2018reinforcement}, Imitation Learning~\citep[IL,][]{hussein2017imitation}, Goal-Conditioned RL~\citep[GCRL, e.g.,][]{schaul2015universal}, Meta-RL~\citep{gupta2018meta}, etc, as a \emph{\lintlong{} (\lint)} setting.

A major challenge for \lint\ methods is task variability in real-world environments~\citep{kormushev2013reinforcement, zhu2020ingredients}. A viable \lint\ agent's policy is expected to do well, out of the box or after a brief adaptation, not only on the specific tasks it encountered during training but also on a multitude of related ones that may differ in dynamics, rewards, initial states, and visual appearance. In other words, a \lint\ agent needs to \emph{generalize} well from its potentially limited experience on a training \emph{task distribution} to a task distribution it faces at deployment time. Unfortunately, the time and cost of training and evaluating agents in real-world environments, including the potential of catastrophic consequences of suboptimal agent behavior~\citep[see, e.g., ][for a comprehensive discussion]{ibarz2021train}, make research on generalization of \lint{} methods directly in these domains prohibitively difficult. Recognizing this, researchers have proposed a number of benchmarks to facilitate \lint\ generalization studies~\citep{packer2018assessing, cobbe2019procgen, song2019observational, minigrid2021, robosuite2020, ahmed2020causalworld, hafner2021benchmarking}. However, none of these environment suites give a researcher the flexibility to easily define a fine-grained \emph{generalization objective} and evaluate a \lint\ method with respect to it. The current lack of accessible tools for evaluating \lint\ agents on carefully constructed generalization objectives limits our ability to establish whether an agent learns spurious correlations between noise and signal~\citep{song2019observational}, truly understands the environment~\citep{ke2021systematic}, can generalize from limited examples~\citep{finn2016generalizing, kahn2018self}, and can perform well on tasks that are drawn out-of-distribution compared to the training ones~\citep{packer2018assessing}.

\begin{figure*}[t!]
    \centering
    \includegraphics[width=\linewidth]{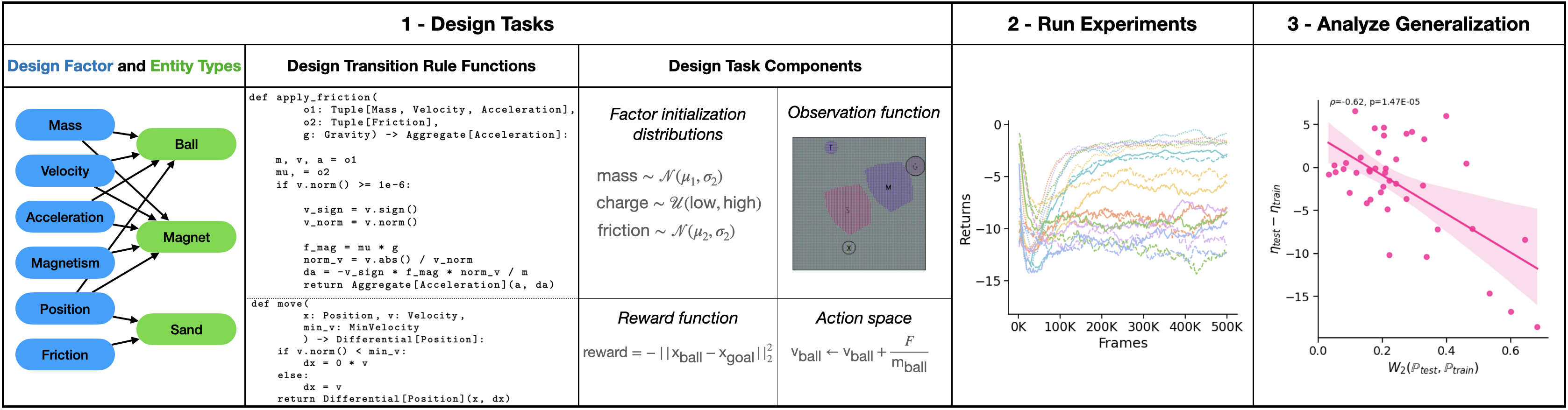}
    \caption{SEGAR at a glance. SEGAR begins with design of tasks, which is done by defining factor and entity types (collections of factor types), then using these types to define rules for the transition function and the other components of the task. After agents are trained on these tasks, this design process enables the researcher to perform detailed analysis on the agents' generalization capabilities.}
    \label{fig:main_story_figure}
\vspace{-.5cm}
\end{figure*}

Our work introduces the \segarlong{} (\segar), an environment designed to fill this research need. \segar, whose high-level schematic is shown in Fig.~\ref{fig:main_story_figure}, is a toolkit for creating tasks, and we provide examples inspired by minigolf and billards. \segar's main distinguishing feature is providing direct access to factors of variation (e.g., object masses, friction, etc) and rules that define tasks in this environment. In particular, \segar\ allows a researcher to design and construct distributions over the values of factors of variation and dynamics-governing rules, which \segar\ automatically translates to distributions over computationally lightweight tasks with user-defined observability properties. The construction process is highly \emph{customizable}: a researcher can choose from pre-defined dynamics rules or write their own. Defining a pair of task distributions in this way amounts to inducing a generalization objective. \segar\ contains tools for quantifying the difference between user-defined task distributions, letting researchers characterize the generalization gap a \lint\ agent is expected to bridge in a given experiment and thereby facilitating accountability in experiment design. Last but not least, as we believe that studying \lint\ agents' \emph{representations}~\citep{bengio2013representation} will be crucial for developing better algorithms, \segar\ provides means of evaluating learned representations w.r.t. a task's underlying state space.
\segar{} is open-sourced at \href{https://github.com/microsoft/segar}{https://github.com/microsoft/segar}. In this paper, we present an overview of \segar{} and show several experiments illustrating a few types of research questions \segar{} can help answer.

\comment{
Unfortunately, the time and cost of training and evaluating agents~\citep{berner2019dota}, the inherent variability of real-world settings~\citep{kormushev2013reinforcement, zhu2020ingredients}, and potential catastrophic consequences in them~\citep[see][for a comprehensive discussion]{ibarz2021train} can make \lint{} research directly in these domains prohibitively difficult.
To make studying LInt more practical, researchers have created artificial environments, e.g., based on video games~\citep{bellemare2013arcade, cobbe2019procgen}, simulated robotic control~\citep{robosuite2020, ahmed2020causalworld}, and autonomous driving~\citep{dosovitskiy2017carla}.
While these environments have enabled progress~\citep{zhu2018reinforcement, zhao2020sim}, less progress has been made with agents that operate well under variation of tasks within an environment, an inevitable feature of real-world settings.
The \emph{task distribution} can describe such variability, evident in environment properties such as the observation space, initial conditions, transition functions, or reward function.
One can also expect in real-world settings that the agent will have a limited budget of training tasks, which can lead to overfitting thus poor performance at deployment, and the seriousness of this can depend on the complexity of the training and test task distributions (e.g., multimodality).
The \emph{objectives} we wish to highlight are those that involve good performance in different set of tasks at deployment than in training, in other words \emph{generalization objectives}.
Not evaluating agents on generalization objectives limits our ability to evaluate whether the agent learned spurious correlations between noise and signal~\citep{song2019observational}, truly understands the environment~\citep{ke2021systematic}, can generalize from limited examples~\citep{finn2016generalizing, kahn2018self}, and can perform well in or adapt to tasks that are drawn out-of-distribution w.r.t. the training ones~\citep{packer2018assessing}.

Thankfully, there has been a growing movement towards environments that facilitate progress towards this common objective~\citep{packer2018assessing, cobbe2019procgen, song2019observational, minigrid2021, robosuite2020, ahmed2020causalworld, hafner2021benchmarking}.
Towards this goal, we developed the \segarlong{} (\segar), which brings several crucial components into one place (see Fig.~\ref{fig:main_story_figure} for a high-level schematic).
SEGAR is designed with generalization in mind by providing \emph{intuitive and transparent control over task distributions}.
In order to facilitate this, the state space is exposed with initial conditions fully specifiable by known parametric distributions.
This allows for \emph{quantitative measures on experimentation}, such as measuring how representative a set of train tasks are of their underlying distribution or how close train and test task distributions are.
This experimental measurability in turn improves accountability in experiment design.
In addition, \segar{} is designed to be highly \emph{customizable}, allowing for extensibility on generalization experimentation.
Finally, as we believe that studying the representation of interactive agents will be crucial towards developing better algorithms, we provide \emph{means of evaluating representations} of the agent w.r.t. the underlying state space.
We note that, while we provide these built-in tasks and measures, \segar{} is designed from the ground up to be customizable and extensible in nearly any way, allowing the researcher to design new tasks with novel components and measures.
\segar{} can be found at \href{https://github.com/microsoft/segar}{https://github.com/microsoft/segar} and is open-sourced under an \href{https://github.com/microsoft/segar/blob/main/LICENSE}{MIT license}.
}

\section{Motivation and related work}

It is standard to use a generalization objective to evaluate a learning algorithm when the data is drawn i.i.d..
Objectives can challenge a model's ability to generalize in many ways, from succeeding despite limited training data~\citep{tan2019efficientnet}, 
to predicting classes not seen during training~\citep[e.g., zero-shot learning][]{xian2018zero}, and to adapting to distributional shifts without forgetting~\citep{zenke2017continual}, to name a few.

Crucially, settings with i.i.d. data have "natural" benchmarks for evaluating generalization that \lintlong{} (\lint) settings don't. Datasets such as ImageNet~\citep{deng2009imagenet} consist of real data, and are believed to be representative of real-world data distributions. Therefore, evaluating generalization of a supervised ML method on just a few benchmarks like this is typically considered sufficient for demonstrating the method's real-world relevance. In contrast, due to the vagaries of evaluation in real-world \lint\ settings, most of the widely used \lint\ benchmarks are synthetic (i.e., potentially not real-world-relevant) and lack fine-grained control over how their task distributions are generated. 
As a result, existing \lint\ benchmarks don't provide a reliable way of evaluating generalization objectives.
Environments such as the Procgen benchmark~\citep{cobbe2019procgen} provide some control over procedural generation of various parameters of the task~\citep{minigrid2021, fan2021secant, kuttler2020nethack, robosuite2020, johnson2016malmo, hafner2021benchmarking, juliani2019ot}, but the true distribution of the task may be obscured by complex control flows.
Beyond this, for standalone discrete-state and -action tasks whose transition functions are expressible in terms of logical predicates, control over task generation is offered by description languages such as PDDL~\citep{pddl} and RDDL~\citep{Sanner_relationaldynamic} and tools such as PDDLGym~\citep{silver2020pddlgym} that compile task descriptions into simulators. 
For task distribution control, various environments provide some control, such as CausalWorld~\citep{ahmed2020causalworld}, MDP playground~\citep{rajan2019mdp}, Textworld~\citep{cote2018textworld}, and Mini-environments~\citep{ke2021systematic}.
Control and transparency over the distributions of tasks are necessary for making convincing conclusions on the generalization performance of agents and their algorithms.
For \segar{}, we take the stance that the more transparent and fine-grained the control of the task distributions an environment provides, the stronger the benchmarks it enables for evaluating generalization, and we implement this idea by making this the starting point in our design.



Given transparent control over task distributions, quantitative measures theoretically become available to reveal the nature and hardness of generalization problem.
Towards this, MDP playground~\citep{rajan2019mdp} enables measuring the task hardness, and XLand~\citep{team2021open} provides an extensive set of metrics on tasks.
As SEGAR provides full access to the distribution that generated the tasks, this allows for direct measurement of distances between task distributions and samples. 

Another crucial design choice of artificial environments is the structure and complexity of the task, the choices of which can have significant impact on experimentation cost and application to real-world settings.
A number of popular environments are video games~\citep{bellemare2013arcade, cobbe2019procgen, team2021open} or grid worlds~\citep{ke2021systematic, minigrid2021} with visual observations and dynamics that have little relevance or resemblance to the real world.
3D video game environments~\citep{team2021open, hafner2021benchmarking, johnson2016malmo} include some more real-world physics (such as gravity), but the resemblances are overall small.
A number of robotic~\citep{robosuite2020, ahmed2020causalworld} and physics~\citep{todorov2012mujoco} environments boast realistic physics with various degrees of customization, but always come with a large trade-off between speed, stability, and accuracy~\citep{erez2015simulation, dulac2019challenges}.
SEGAR strikes a balance between realism and cost through flexibility, allowing the researcher to structure the task in a way that reflects what they believe represents real-world settings and to choose where to pay the cost of realism.
There is also a substantial gap in the above features in 2D environments, as most of those that fulfill the above criteria are in 3D.
Taking inspiration from Box2D~\citep{box2d}, SEGAR fills this gap, facilitating the movement from strong SOTA algorithms in 2D environments to more complex 3D ones~\citep{kolve2017ai2, savva2019habitat}, where strong algorithms are in short supply.
Finally, SEGAR is a sandbox environment, allowing for varying nearly every aspect of the task, from the observation and state spaces to the dynamics and reward functions.


A widely studied class of approaches to building an agent or a model whose behavior generalizes to new circumstances is encouraging the agent to learn a latent \emph{representation} of its inputs. 
There are a number of RL methods that use a representation learning loss function to improve performance, whether related to generalization or not~\citep{laskin2020curl, mazoure2020deep, schwarzer2020data, schwarzer2021pretraining, hafner2020mastering, mazoure2021cross, yarats2021reinforcement}.
However, despite the loss functions operating directly on the agent's representation, the representations themselves are rarely evaluated beyond evaluating the returns, with some exceptions~\citep{such2018atari, anand2019unsupervised, zhang2020learning, mazoure2021cross, lyle2021effect}.
However, if the underlying factors of the environment are known, it can be straightforward to test where representational properties correspond to success on generalization objectives.
As \segar{} provides fine-grained control over distributions of factors, we can intervene on these distributions and study the effect on the resulting representations as well as downstream performance.


\section{Terminology and setting definition}
\segar{} is a sandbox environment and toolkit for defining generalization objectives via distributions of sequential decision-making \emph{tasks}, automatically translating them to distributions of \emph{partially observable Markov decision processes (POMDPs)}.
\segar{}'s tasks involve objects (\emph{entities}) and \emph{rules} that determine how objects interact with each other depending on the objects' properties (\emph{factors}). \segar{} allows for easily defining distributions over factors' values, and thereby over tasks.
We formalize all the highlighted terminology in the Supp. for describing \segar{}'s tools in \Cref{sec:segar}.

\segar{} compiles every task into a mathematical formalism widely adopted in IL and RL, a \textbf{POMDP}. For a task $T$, POMDP $M_T$ is a tuple $\langle \mathcal{X}, \mathcal{A}, T, r, \mathcal{O}, z, x_0\rangle$, where the state space $\mathcal{X}$, the action space, $\mathcal{A}$, the reward, $r$, the observation function, $z$ are defined as above, $\mathcal{O}$ is the observation space induced by $z$, and $P: \mathcal{X} \times \mathcal{A} \mapsto \Delta(\mathcal{X})$ is a transition function induced by $T$'s rule set $\mathscr{R}$.
The mechanisms for combining rules into a transition function will be covered in~\cref{sec:segar}. 
Finally, $x_0$ is the POMDP's initial state, derived from the factor values from $T$'s initial configuration $c_0$. 
An optimal POMDP solution is a \textbf{policy} $\pi:\mathcal{H} \mapsto \Delta(\mathcal {A})$ that maps histories of observations $h_t \in \mathcal{H}, h_t = o_t, o_{t-1}, \ldots, o_1$ to action distributions so as to maximize some objective, e.g., expected discounted sum of rewards over a distribution of tasks.


\comment{
========================== OLD VERSION ==========================
We define a \emph{parametric task} $T_\theta$ as a partially observable Markov decision process (POMDP) $T_\theta=\langle \cS,p_{0;\theta},\cA,p_{\cS;\theta},\cO,p_{\cO;\theta}, r, \gamma \rangle$, where $\cS$ is a state space, $p_{0;\theta}=\mathbb{P}[s_0;\theta]$ is the starting state distribution, $\cA$ is an action space, $p_{\cS;\theta}=\mathbb{P}[\cdot|s_t,a_t;\theta]:\cS\times \cA \to \Delta(\cS)$ is a transition function, $\cO$ is an observation space, $p_{\cO}=\mathbb{P}[\cdot|s_t;\theta]:\cS \to \Delta(\cO)$\footnote{$\Delta(\cX)$ denotes the entire set of distributions over the space $\cX$.} is an observation function, $r:\cS\times \cA \to \mathbb{R}$ is a reward function and $\gamma\in [0,1)$ is a discount factor. The system starts in one of the initial states $s_0\sim p_{0;\theta}$ with observation $o_0\sim p_{\cO}(\cdot | s_0;\theta)$. At every timestep $t=1,2,3,..$, the agent, parameterized by a policy $\pi:\cO \to \Delta(\cA)$, samples an action $a_t\sim \pi(\cdot|o_t)$. The environment transitions into a next state $s_{t+1} \sim p_{\cS}(\cdot|s_t,a_t;\theta)$ and emits a reward $r_t=r(s_t,a_t)$ along with a next observation $o_{t+1}\sim p_{\cO}(\cdot|s_{t+1})$.

The state space $\cS$ is partitioned into sets of \emph{factors}, i.e. parameters of entities present in the POMDP. Examples of entities include agent, enemies, obstacles, walls, interactive objects, etc, and completely characterize the dynamics occurring inside the environment. 
Each task $T_\theta$ is instantiated by sampling $\theta$ from a task distribution $\mathbb{P}_\theta\in \Delta(\Theta)$ with support $\Theta$.
}

\section{The Sandbox Environment for Generalizable Agent Research \label{sec:segar}}
This section provides an overview on how \segar{} is structured as a programming interface for the goal of designing generalization objectives for \lint{} generalization research. 
\segar{} is functionally split two parts: an environment and a set of experimentation and representation \emph{metrics}. 
The environment is composed of:
\begin{itemize}
    \item A \emph{state space}, $\mathcal{X}$ is composed of all factor type vectors of the entities in the task, and which is determined after the numbers and types of entities have been sampled by the environment.
    \item A \textsc{Simulator} organizes the entities, factors, and rules, executing a \emph{transition function}, $P$.
\end{itemize}
The environment generates tasks, $T$, which compiles into a \textsc{MDP} with the following components:
\begin{itemize}
    \item An \textsc{Observation} object defines the function, $z$, between states and observations, as well as which factors are observable. 
    This could also include a renderer for pixel-based observations, which operates as a function between states and visual features.
    \item A \textsc{Task} object, which is composed of:
    \begin{itemize}
        \item An \textsc{Initialization} object determines the initial state, $s_0$, by sampling the number of and type of entities as well as their initial configurations, $c_0$.
        \item An \emph{action space}, $\mathcal{A}$, determines how the agent can change the underlying states.
        \item A \emph{reward function}, $r$,  defines the returns.
    \end{itemize}
\end{itemize}

All of these components are encapsulated in an \textsc{MDP} object, which is implemented as a OpenAI Gym environment~\citep{brockman2016openai}.
In the next sections, we will present the above components of the environment and describe how each can be used for generalization experiments.


\subsection{State space}
\begin{figure}
    \centering
    \begin{subfigure}[c]{0.2\textwidth}
        \centering
        \includegraphics[width=\textwidth]{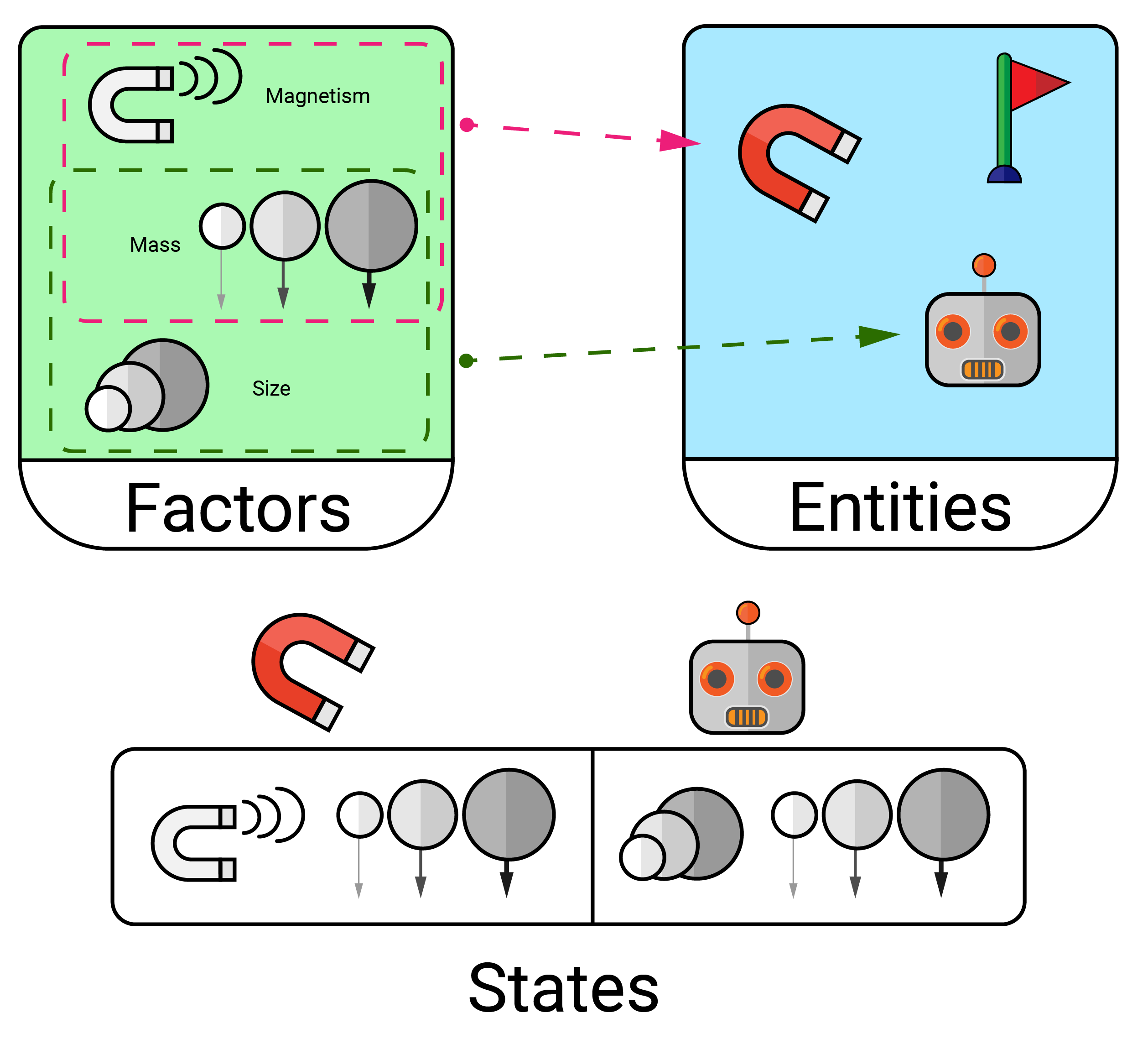}
        \caption{States are partitioned by entity instances, then factor types.}
        \label{fig:state_space}
    \end{subfigure}
    \hfill
    \begin{subfigure}[c]{0.38\textwidth}
        \centering
        \includegraphics[width=\textwidth]{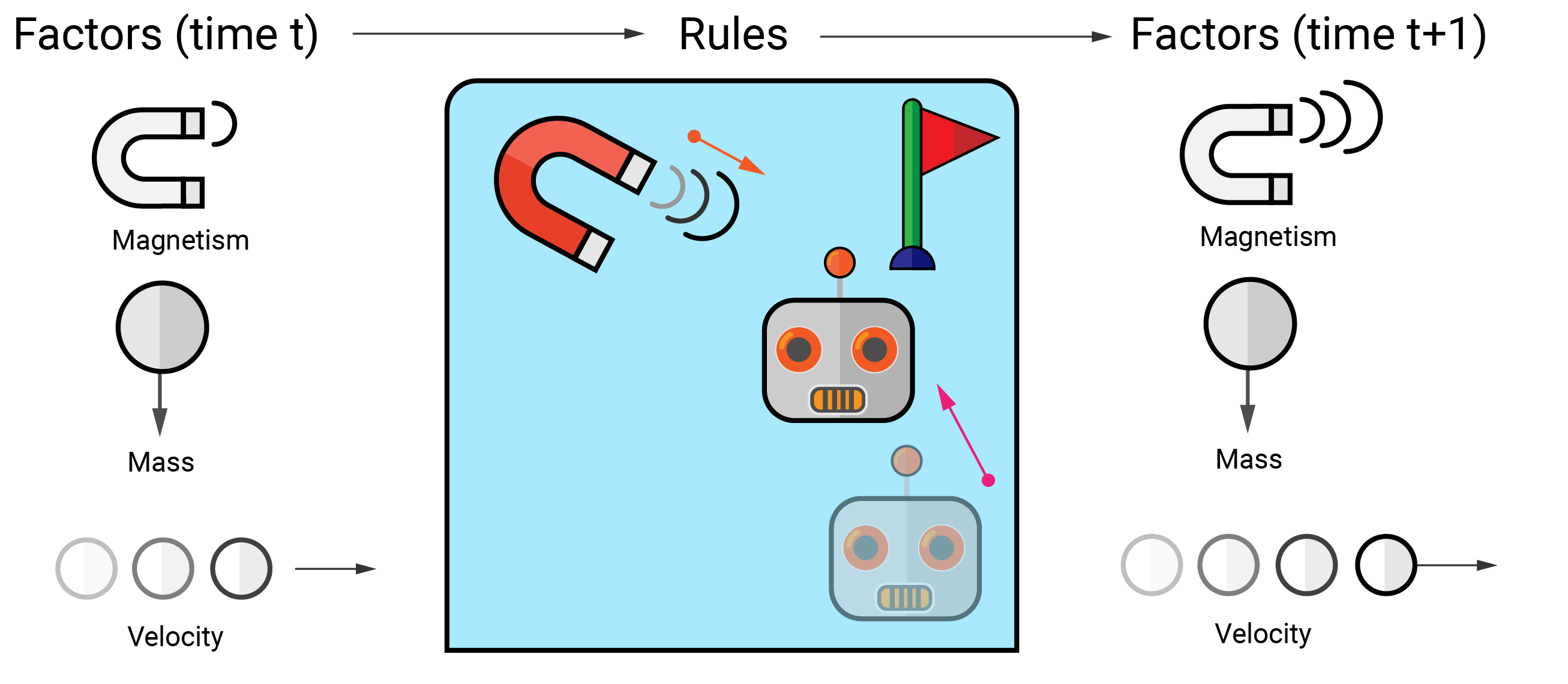}
        \caption{Rules are simple Python functions that change factor values, using the Python typing system. See Fig.~\ref{fig:main_story_figure} or the supplementary material for examples.}
        \label{fig:arule}
    \end{subfigure}
    \hfill
    \begin{subfigure}[c]{0.38\textwidth}
      \centering
      \includegraphics[width=.45\textwidth]{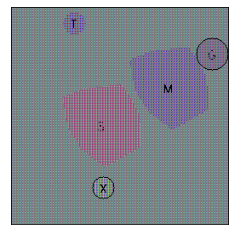}
      \includegraphics[width=.45\textwidth]{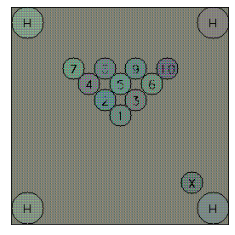}
      \caption{Example renderings of tasks for pixel-based observations. The researcher can use any observation function they wish, and renderers are customizable as well.}
      \label{fig:tasks}
    \end{subfigure}
    \caption{
    The state space is a product of two vector spaces, each corresponding to an entity. 
        Each entity contains a different set of factor types, which make up the basis of each entity vector space, and the state space is not structured beyond being a vector space and the basis ordering fixed by the environment, and it can be of arbitrary size, depending on the numbers and types of entities.
}
\vspace{-.5cm}
\end{figure}

An illustration of the state space is provided in Fig.~\ref{fig:state_space}.
The \segar{} states are composed of a set of \emph{entities}, and the entities are composed of a set of \emph{factors}.
For each task, the \textsc{Initialization} object (see below) samples the number of a given set of entity types, $\mathcal{T}_{entity}$, resulting in a finite multiset of entity types.
Each entity type, $\mathcal{T}^e \in \mathcal{T}_{entity}$, has a unique set of factor types, $\mathscr{T}^e_{factor} \subseteq \mathscr{T}_{factor}$.
The entity types are all subclasses of the \textsc{Entity} generic class, and their factor types are implemented as a set of dictionary keys, each key being a subclass of the \textsc{Factor} generic class.
Some example of types are provided in the Supp., and we provide a full list of built-in types in the \href{https://segar.readthedocs.io/en/latest/autoapi/segar/things/}{documentation}.
Note that all types are extensible through sub-classing any existing type (including the generic types).
 

\subsection{Simulator, rules, and the transition function}
The \textsc{Simulator} controls the environment: it organizes the rules, entities, and factors, applies the appropriate set of rules to each entity, and manages collisions.
The collection of rule applications along with collision dynamics make up the overall \emph{transition function}, $P$, of the POMDP.

\segar{} allows the user to specify the set of rules, and the simulator applies those rules to all of the entities active in the arena of the task (see Fig.~\ref{fig:arule} for an illustration).
Which entities apply to which rule is determined by pattern matching using the Python type hints system ( \href{https://docs.python.org/3/library/typing.html}{https://docs.python.org/3/library/typing.html}).
In order to handle multiple rules applying to potentially the same entity's factors, we implemented a simple logic managed by the simulator to decide what rules to apply (resolve rule conflicts) and how (if at all) their results combine.
Example rules and details on the output types are provided in the Supp.

\subsection{Observations and Renderings}
\segar{} allows the user to chose how the agent sees the underlying states.
The user can specify which factors to include in the observation space, allowing full flexibility in defining an MDP, POMDP, or block MDP.
The \emph{observations} in \segar{} are implemented as a callable \textsc{Observation} object, which are passed to the \textsc{MDP} object for training the agent.
Built-in classes include full state-based observations (all entities), partial state-based observations (some of the entities), 2D pixel-based observations, and multi-modal observations built from any combination of these. 
Details on a the built-in renderers for pixel-based observations are provided in the Supp.

Though we provide built-in renderers for researchers to use, we believe that providing observations to the agent based on visual features that are understandable by humans may unintentionally introduce experimental challenges, as neural networks can leverage low-level visual cues to "cheat" at tasks that were intended to require higher-level reasoning. 
In \segar{}, the visual features are treated as a transparent and controllable component in building experiments, allowing for researchers to design their own observation spaces and renderers to better reflect the domain they wish to study.

\subsection{The Task}

The final component in \segar{} is the \textsc{Task}, which encapsulates the \textsc{Initialization} object, the reward function, and the action space, as well as points to the \textsc{Simulator} and \textsc{Observation} objects.
\segar{} comes with three built-in demonstration tasks: \textsc{PuttPutt}, \textsc{Billiards}, and \textsc{Invisiball}, though \segar{} was designed to easily allow users to define new tasks within the framework.

\begin{itemize}
    \item \textsc{PuttPutt}: Navigate a ball around various obstacles to a goal location.
    \item \textsc{Billiards}: Controlling a cueball, use collisions to knock the other balls into holes.
    \item \textsc{Invisiball}: PuttPutt, but the ball is invisible after the first step. The ball has charge and there are other objects that also have charge, so the agent needs to infer charge and position of the ball through the movement of the other objects.
\end{itemize}

\paragraph{Initialization}
\segar{} allows for defining train and test task distributions for generalization experiments. 
Such experiments are a major motivation behind \segar{}'s design, and \segar{} makes them easy to set up by specifying distributions (such as Gaussian, Uniform, etc) over factors' initial values for train and test tasks.
The \textsc{Initialization} object controls which entity types are initialized and what their initial factor values are, ultimately sampling the initial state, $s_0$.
The \textsc{Initialization} object allows for specification of the initial entity configuration, $c_0$ through distributions, and different initial conditions can be sampled to generate multiple tasks from the same \textsc{Initialization} object.
The parameters of these distributions are specified through a \textsc{Prior}, which is a special type of rule that samples factor values from given parametric distributions.

\paragraph{Reward function and action space} Reward functions defines what success means for the agent in an environment, and \segar{} allows an researcher to easily construct them from the underlying states and the the agent's actions.
The reward function is implemented as a simple Python function that takes the states and actions as input and outputs a scalar.
The action space is the combination of a Gym space and a function which operates directly on the factors of the environment.
For example, a ``force" action on a ball would change the velocity factor of the ball instantaneously.
Valid actions are determined by a function that interacts directly with the entity factors through the simulator, so a designer is not limited on what factors the agent can intervene on through its actions.

\subsection{Metrics on experimentation and representations}
The second part of \segar{} are metrics, both on objectives and representations.
The former is in place to add accountability to design: without metrics on \lint{} tasks, it is difficult to assert that generalization ``success" could -- or even should -- imply success in other settings, notably real-world ones.
This is made possible as \segar{} allows the researcher to define the MDP in terms of distributions.
In addition to providing control over and access to these distributions, \segar{} provides access to statistics on these distributions, such as CDFs, PDFs, entropy, etc.
Using these statistics, we can easy derive the entropy of the initialization, as well as measure how similar two distributions of initialization are.
In addition, we can measure how representative a set of samples, say a training set, are to the underlying distribution they come from using statistics such as the Kolmogorov–Smirnov test~\citep{smirnov1948table}, or even from training samples to the test distribution.
Finally, we can use the Wasserstein-2 distance~\citep{ruschendorf1985wasserstein, flamary2021pot} to compute distances between sets of tasks (say train vs test) through their initializations.
Note that there are a number of proposed ways to measure distances between MDPs~\citep{agarwal2021contrastive}, and as \segar{} provides full exposure to the underlying factors and their distributions, we look forward to their further development and inclusion.

For representation evaluation, we have full access to the underlying state space, so we can use mutual information as a measure for the agent's understanding of the state space.
We use Mutual Information Neural Estimation~\citep[MINE,][]{belghazi2018mutual} but use a less biased version based on the Jensen-Shannon divergence~\citep{hjelm2018learning, poole2019variational}.
As this amounts to training a classifier, we can provide accuracy as a score for how well the agent understands the underlying factors.
These metrics are explored in the following sections as examples of analysis possible using \segar{}.

\section{Experiments}


\segar{} is primarily a tool which can answer important questions regarding generalization of learning agents and their world representations. 
Some of these questions include, \emph{``How does the performance of an agent trained on task set A and tested on task set B correlate with the distance between sets A and B?"} and \emph{``Does an agent need to properly identify the various latent factors of an environment such as position, velocity, mass, and gravity for better generalization performance?"}. 
We attempt to provide an answer to some sample questions in a controlled experimental setup, defined below.



\paragraph{Experimental setup.} Let $\mathbb{P}_{train}$ and $\mathbb{P}_{test}$ be two task distributions on the same support. Since, in practice, their distribution functions are unknown, the learner has only access to sample tasks $T_1,..,T_n\sim \mathbb{P}_{train}$ and $T'_1,..,T'_m\sim \mathbb{P}_{test}$, and their corresponding sets of parameters. We first form an epistemic POMDP~\citep{ghosh2021generalization} out of $T_1,...,T_n$, where every episode is samples from one of the $n$ tasks uniformly at random. We then fit a PPO agent~\citep{schulman2017proximal} on this new task, and train it until convergence of returns, or exactly 1M steps. After that, the agent's policy and value networks are snapshotted, and can be re-used in all representation learning probes described later in this section. 
In addition, we include Pearson's $\rho$ correlation and the corresponding $p$-value for plots where linear trend lines are fitted.
For all plots, we treated points with a standard $Z$-score of $3$ or greater as outliers and removed them from our analysis.

\begin{figure*}[h!]
    \centering
    \begin{subfigure}[b]{0.24\textwidth}
    \centering
    \includegraphics[width=\textwidth]{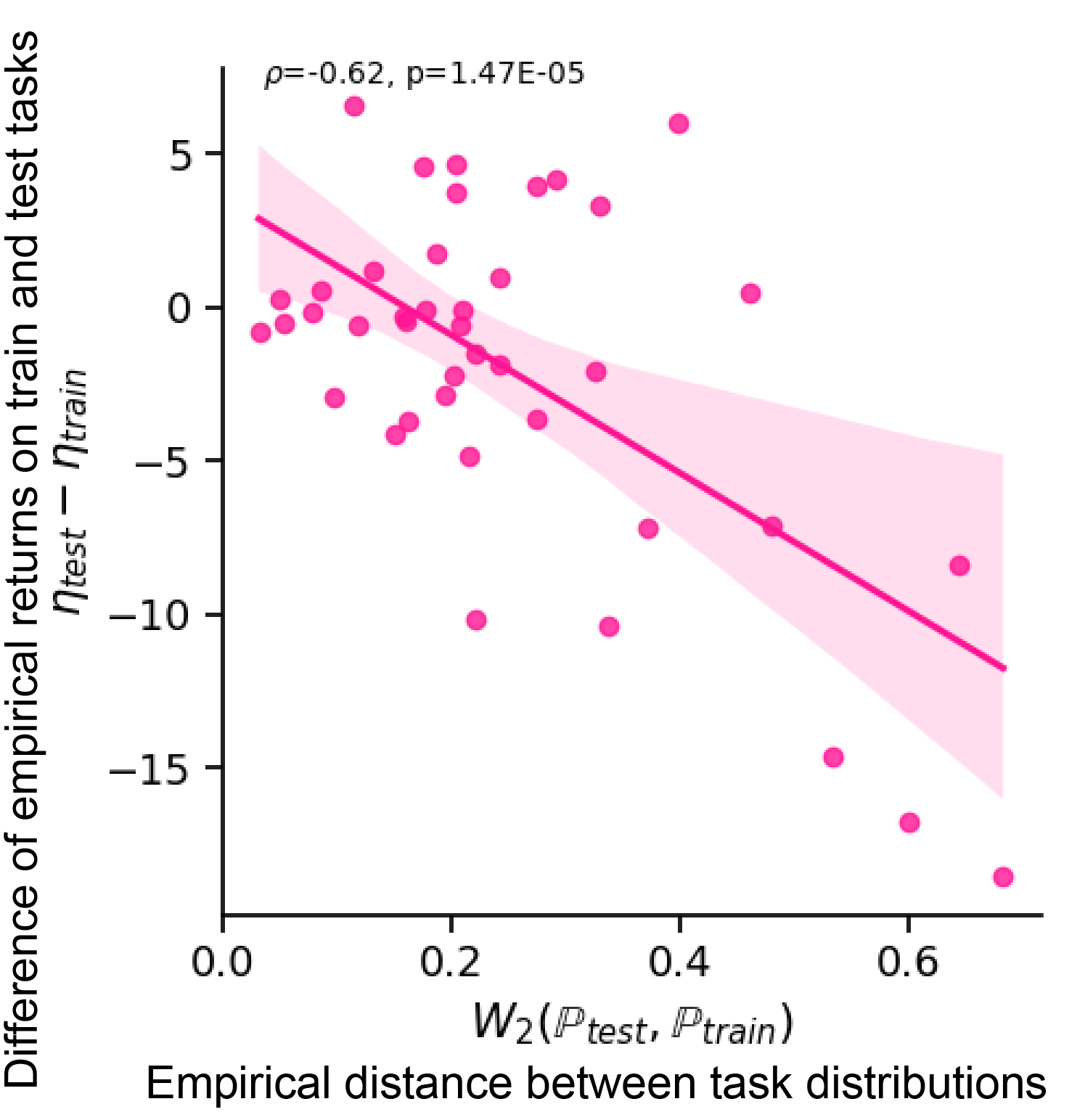}
    \caption{}
    \label{fig:returns_vs_w2}
    \end{subfigure}
    \hfill
    \begin{subfigure}[b]{0.74\textwidth}
    \centering
    \includegraphics[width=\textwidth]{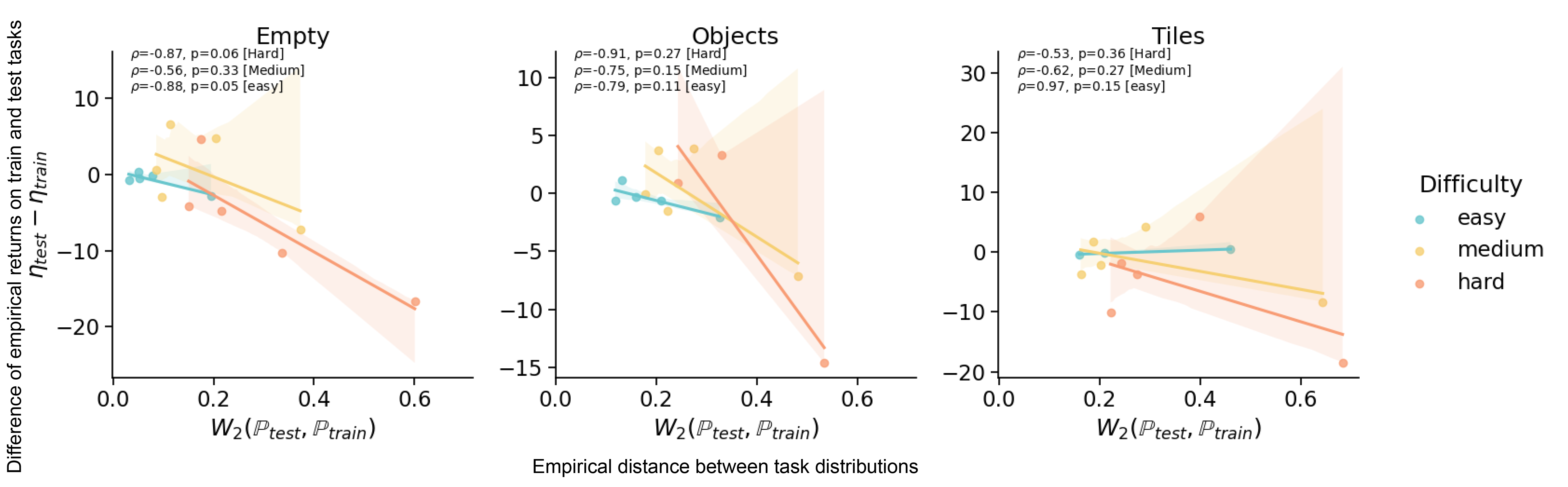}
    \caption{}
    \label{fig:returns_vs_w2_conditional}
    \end{subfigure}
    \caption{
    (a) Generalization performance gap of a PPO agent as a function of the sample-based Wasserstein-2 distance between latent factors of $n$ random train tasks and 100 test tasks, aggregated by task type and difficulty.
    (b) Generalization performance gap of a PPO agent as a function of the sample-based Wasserstein-2 distance between latent factors of $n$ random train tasks and 100 test tasks, split by task type and difficulty. Each point represents the performance of an agent trained on a different number of levels $n$. Trend lines show the least-squares line of best fit, together with 95\% confidence intervals, Spearman $\rho$ coefficient and $p$-values.}
\vspace{-.3cm}
\end{figure*}

\begin{figure*}[ht!]
\centering
\begin{subfigure}[t]{0.74\textwidth}
    \centering
    \includegraphics[width=\linewidth]{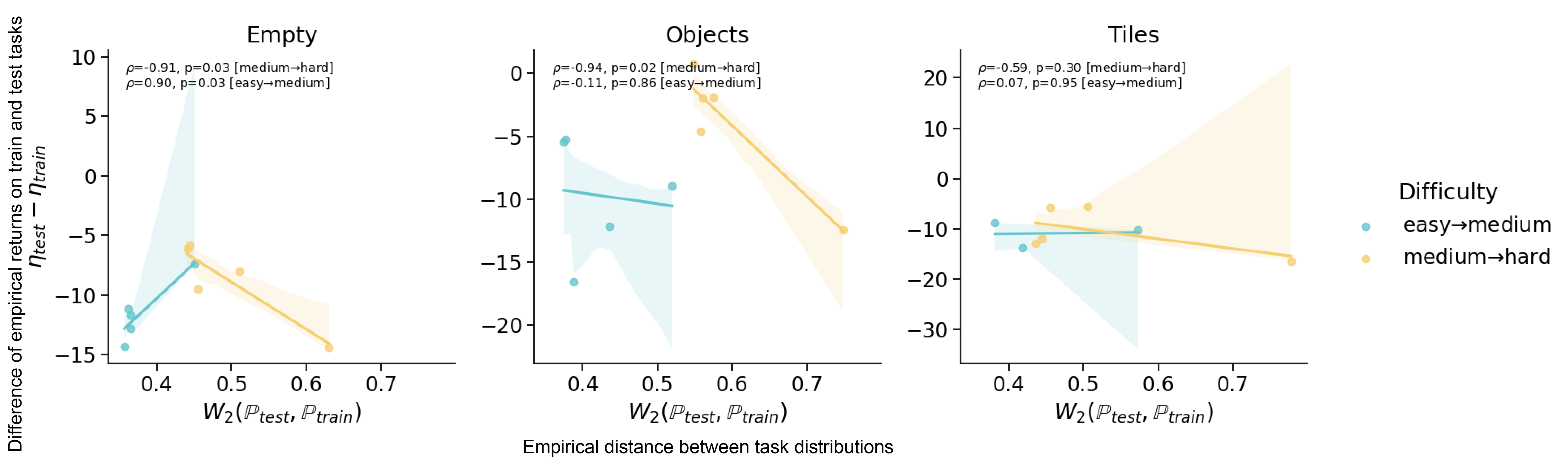}
    \caption{}
    \label{fig:returns_vs_w2_conditional_ood}
\end{subfigure}
\begin{subfigure}[t]{0.24\textwidth}
    \centering
    \includegraphics[width=1\linewidth]{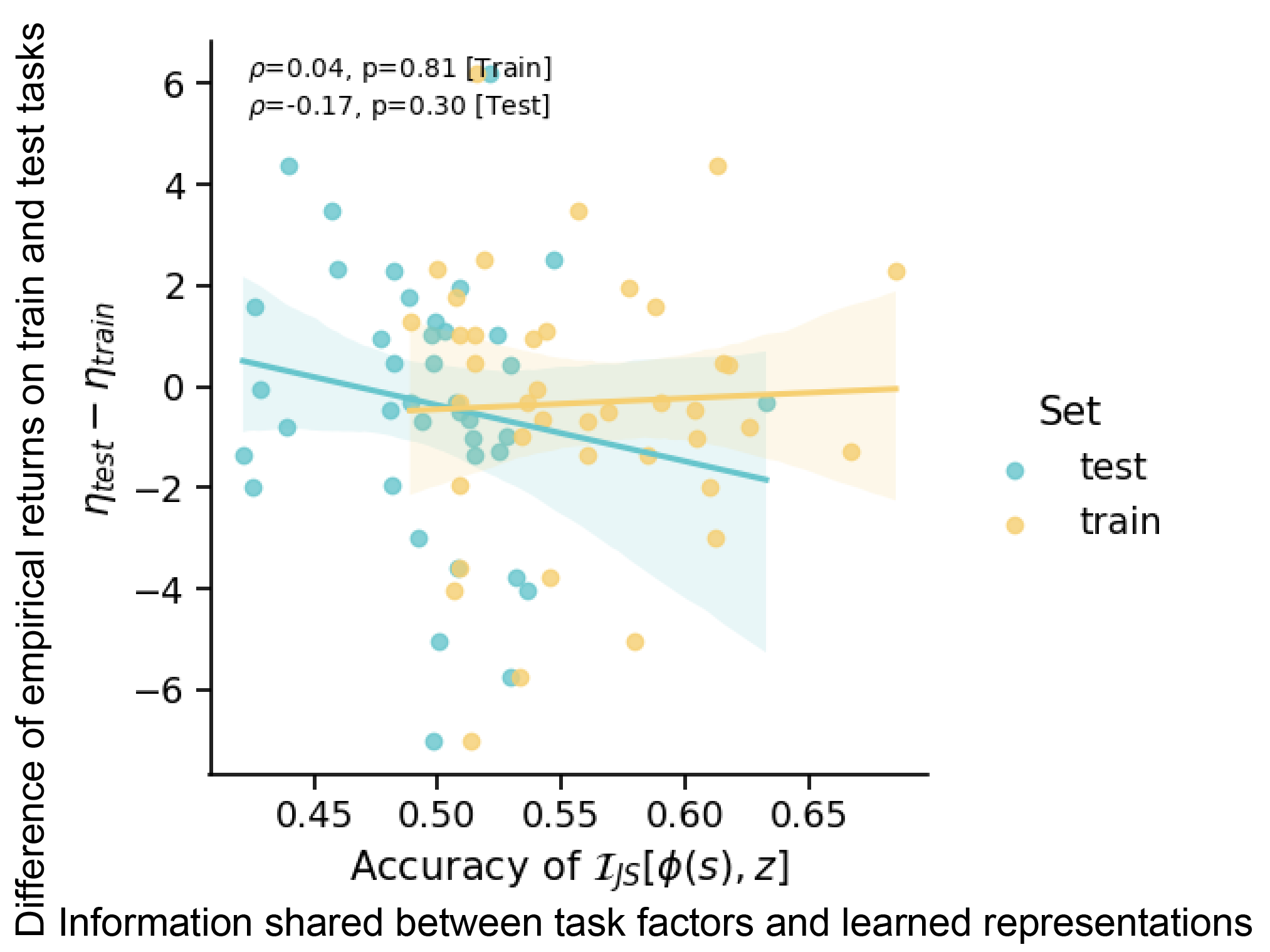}
    \caption{}
    \label{fig:returns_vs_mine}
\end{subfigure}
\caption{
(a) Generalization performance gap of a PPO agent as a function of the sample-based Wasserstein-2 distance between latent factors of $n$ random train tasks and 100 test tasks, split by task type and difficulty. The test distribution has a higher difficulty level than the corresponding training distribution (easy $\to$ medium, medium $\to$ hard). \textit{Generalization gap increases with higher $W_2$ values.} (b) Generalization performance gap of a PPO agent as a function of the lower-bound on mutual information between learned state representations and latent environment factors. Each point represents the performance of an agent trained on a different number of levels $n$. Trend lines show the least-squares line of best fit, together with 95\% confidence intervals, Spearman $\rho$ coefficient and $p$-values. \textit{Smaller generalization gap doesn't necessarily require better information on latent factors.}}
\end{figure*}

\paragraph{Does factor distance correlate with generalization gap?}
How does an agent trained only on experiences from $T_1,..,T_n$ re-use this knowledge to solve $T'_1,..,T'_m$? Intuitively, the more $\mathbb{P}_{train}$ and $\mathbb{P}_{test}$ overlap, the more information about one can be used to solve the other. We tested this hypothesis by jointly training an agent on $T_1,..,T_n$, and subsequently measuring it's average performance $\eta_{train}$ on $\mathbb{P}_{train}$ and $\eta_{test}$ on $\mathbb{P}_{test}$ via independent rollouts. Then, the pairwise Wasserstein-2 distance between two tasks $T_1,T_2$ can be computed by solving the classical optimal transport problem with a Euclidean cost between factor values (pseudocode provided in the Supp.).
Both the performance and Wasserstein gaps are shown in Fig.~\ref{fig:returns_vs_w2}, which indeed hows a significant correlation ($p=\num{1.47e-5}$) between better performance on unseen tasks and similarity of these tasks to the training samples.
Additionally, Fig.~\ref{fig:returns_vs_w2_conditional} breaks down the effects between various environment types: those with only agent and the goal entities, those with objects (such as magnets) and with tiles (such as sand), as well as by difficulty levels. The difficulty levels are regulated by the entropy of the task distribution $\mathbb{P}_{train}$. 
The trend of the generalization gap clearly worsens for all tasks as the Wasserstein-2 distance increases, with exception to the easiest tasks, where there is no clear trend.
However, this trend is not as clear beyond the ``Empty" level, as ``Objects" and ``Tiles" levels are weakly significant at best.
Finally, Fig.~\ref{fig:returns_vs_w2_conditional_ood} shows the decrease in performance which happens when transferring a pre-trained agent onto a harder set of tasks (easy$\to$medium and medium$\to$hard, respectively). In this setting, the performance gap is considerably larger than the one in Fig~\ref{fig:returns_vs_w2_conditional}, since the distribution of test tasks is harder.
Similar to previous results, there is a trend for the gap to worsen as the distance between train and test increases for medium$\to$hard, at least with ``Empty" and ``Objects".

\paragraph{Does mutual information between state representations and factors correlate with generalization gap?}
So far, we've identified some correlations between the generalization gap and factor distance between tasks. 
Yet, nothing was stated about the internal world representation that the agent adopts from task samples $T_1,...,T_n$. 
We conducted a second set of experiments to probe the state representation that an agent has about $\mathbb{P}_{train}$ and $\mathbb{P}_{test}$ by only having access to $T_1,...,T_n$. 
To do so, we first trained a PPO agent on $T_1,...,T_n$ until convergence. 
The trained encoder was then used to map pixel observations onto latent representation vectors. 
We then used the MINE estimator~\citep{belghazi2018mutual} parameterized by a simple two-layer MLP to measure the lower-bound on mutual information between these latent representations and the entire set of factors associated with that observation; we denote this lower-bound based on the Jensen-Shannon divergence as $\mathbb{E}_{s,z\sim \mathbb{P}^\pi}[\mathcal{I}_{JS}(\phi(s),z)]$, as outlined in~\citet{poole2019variational}.
Rather than reporting the mutual information, which is unbounded, we report the classification accuracy of the MINE estimator on the binary decision task, which is bounded between $0$ and $1$.
Fig.~\ref{fig:returns_vs_mine} shows the performance of an RL agent as a function of the lower-bound on mutual information between learned state representations and environment factors. We train on an increasing number of task samples, while testing on 500 tasks from the corresponding training distribution.
While the train tasks score higher than the test, there is no relationship between performance and our mutual information estimates, at least with PPO.


\begin{figure}[ht!]
    \centering
    \begin{subfigure}[b]{0.36\textwidth}
    \centering
    \includegraphics[width=\textwidth]{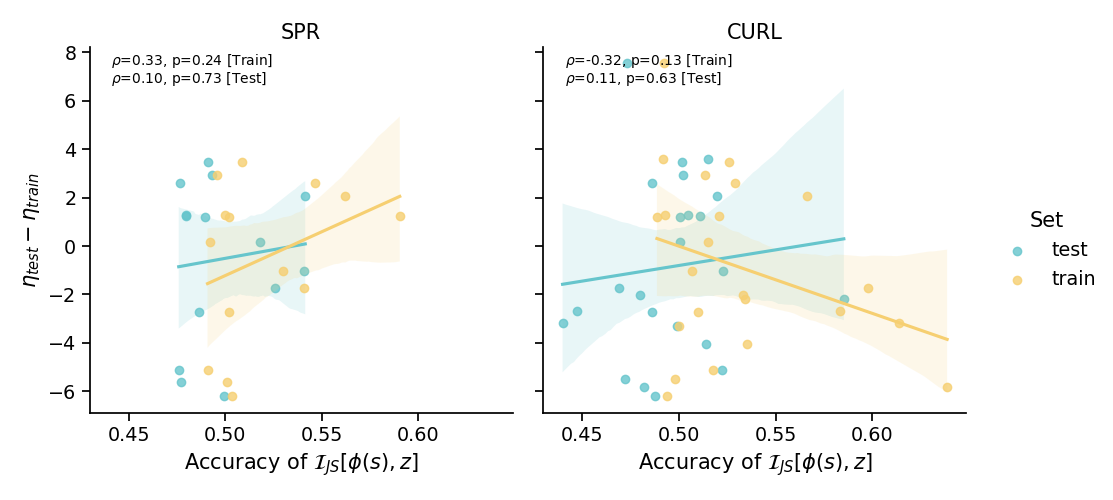}
    \caption{}
    \label{fig:returns_vs_mine_repl}
    \end{subfigure}
    \begin{subfigure}[b]{0.6\textwidth}
    \centering
    \includegraphics[width=\textwidth]{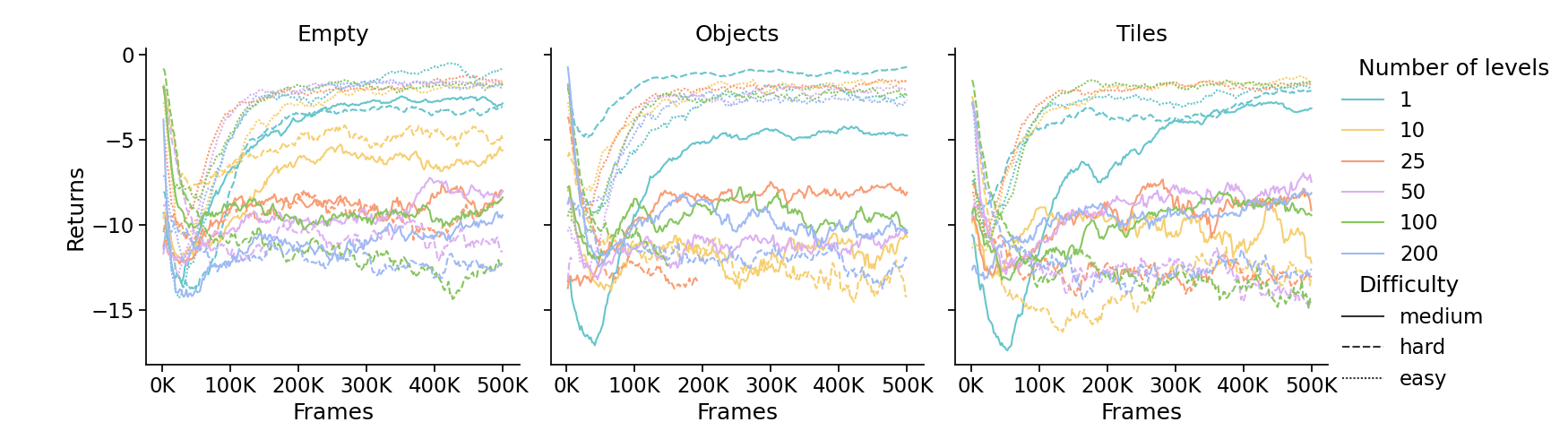}
    \caption{}
    \label{fig:returns_vs_frames_train}
    \end{subfigure}
    \caption{
    (a) Generalization performance gap of a PPO agent as a function of the mutual information between learned state representations and latent environment factors, when adding popular self-supervised objectives. Each point represents the performance of an agent trained on a different number of levels $n$. Trend lines show the least-squares line of best fit, together with 95\% confidence intervals, Spearman $\rho$ coefficient and $p$-values. \textit{Self-supervised signals can improve information about latent factors, and the generalization gap}.
    (b) Performance of a PPO agent as a function of training frames. Each curve corresponds to a different task, difficulty and number of levels configuration. \textit{Agent's performance depends mostly on diversity of training tasks.}
    }
\end{figure}

\paragraph{Do self-supervised representation learning objectives help extract factor information?}

In the past years, unsupervised and self-supervised representation learning methods have been empirically shown to significantly improve generalization capabilities of classical RL agents~\citep{laskin2020curl,mazoure2020deep,schwarzer2020data,mazoure2021cross}. But do these popular representation learning methods improve the understanding of the RL agent of the underlying latent factors?
Fig.~\ref{fig:returns_vs_mine_repl} shows the generalization performance gap as a function of accuracy derived from the mutual information between states and factors for two such popular self-supervised learning~(SSL) methods, CURL~\citep{laskin2020curl} and SPR~\citep{schwarzer2020data}. 
SPR, which relies on predicting exponentially-averaged copies of previous representations, appears to obtain a better generalization gap with higher information content, but only at training.
Evaluation performance appears to be similar for both SSL methods as compared to vanilla PPO.
However, as both SPR and CURL rely on data augmentations, it could be that the augmentations used for the original baselines they were evaluated are ill-suited for other observation spaces.
Finally, as the JSD-based mutual information estimator is high variance~\citep{poole2019variational}, none of the apparent trends are significant, so either more samples or a lower-variance estimator should be used.

\paragraph{Does the RL agent learn well-performing policies?}
Finally, Fig.~\ref{fig:returns_vs_frames_train} shows the learning performance of various PPO agents as a function of training samples, on different environments, difficulty levels and number of task samples.
We can see that the learning is fairly sample-efficient for easier configurations of the tasks, while harder tasks (e.g. with wider variations, objects and tiles) require the agent to train on much more samples before seeing reasonable performance.

\section{Discussion}
The \segarlong{} (\segar) is an open source sandbox environment designed for generalization research in settings with interaction data, e.g., \lintlong{} (\lint).
We demonstrated some of ways we envisioned \segar{} could be used, but due to limited time and resources we were not able to demonstrate nor test everything we had in mind.

\paragraph{Optimal policies and expanding task distribution configurations and benchmarks}
We selected only a few possible configurations of distributions of tasks, labeling them according to the known properties of distributions, such as entropy, and the numbers of and types of objects in the scene.
We limited these configurations based on the performance of PPO, which due to time and compute constraints is the main RL algorithm we explored.
Other baselines exist that could solve tasks within \segar{} much better than PPO~\citep[e.g.,][]{hafner2020mastering}, and testing these stronger baselines could substantially change the picture as far as what is solvable under \segar{}'s full configuration capabilities.
This, in turn, would lead to a larger set of more interesting generalization objective benchmarks.
What follows is a set of potential examples, which are non-exhaustive.

\paragraph{Quantitative measure of hardness based on the reward}
The reward function in \segar{} is flexible and can be clearly related to the underlying factors.
For example, a PuttPutt task could be made more complex by conditioning the reward on more factors or events, such as only giving reward for reaching the goal only after it has also ``rolled" over sand, or require one or more collisions in the process of reaching the goal.
A researcher could use intuitive reward design to scale the hardness, measuring how different algorithms respond to this scaling.

\paragraph{More comprehensive measures on tasks and MDPs}
Measuring tasks based on the initial state or the distribution those states were drawn from are admittedly a small part of the full task, as the full task involves the dynamics due to the transition function as well as the reward.
A key difficulty is that task diversity in terms of dynamics is entangled with the agent's policy, though there are measures between sets of task assuming the optimal policy~\citep{agarwal2021contrastive} that are relevant in transfer learning settings.
Other measures that could separate the exogenous versus endogenous components of the MDP~\citep{dietterich2018discovering} to possibly measure how controllable a set of tasks are as well as use the uncontrollable components to build a policy-agnostic measures. 
These sorts of measures could provide a more comprehensive look of how a set of tasks vary or how different two sets of tasks are.

\paragraph{Varying rules of the environment}
\segar{} is fully capable of providing generalization objectives that involve when the rules change, as this can be done by conditioning the rules on abstract entities whose role is only to provide abstract variable factors that modify those rules, such as the gravitational constant.
One interesting direction would be to explore how variation in the rules effect performance of standard baselines, as well as how agents respond to distributional shifts.


\bibliographystyle{icml2022}
\bibliography{bibliography}
\section{Appendix}
\renewcommand{\thesection}{\Alph{section}}
\renewcommand{\thesubsection}{\Alph{section}.\arabic{subsection}}
\renewcommand{\thesubsubsection}{\Alph{section}.\arabic{subsection}.\arabic{subsubsection}}
\setcounter{section}{0}
\section{Formal definitions of SEGAR's variables}
\label{sec:formal}
Here, we formalize the components of tasks as defined by \segarlong{} (\segar{}).

Task definitions are founded on the concept of \textbf{factor type}. Factor types are physical properties of objects, e.g., \textsc{Mass}, \textsc{Charge}, and \textsc{Position}. 
Each factor type has a base type in its definition, such as float, boolean, int, and can have a base type from a Python library, such as \href{https://numpy.org}{numpy}~\citep{harris2020array} arrays to represent vector-like factor types.
\segar{} provides a number of predefined factor types inspired by real-world physics and allows researchers to add new ones. 
We denote the set of all factor types available to an researcher as $\mathscr{T}_{\mathit{factor}}$. 

Factor types give rise to the concept of \textbf{entity type}. 
Entity types are types of objects in \segar{}'s tasks, such as \textsc{Sand} or \textsc{Magnet}. 
As with factor types, \segar{} provides several built-in entity types and lets the researcher add more. 
An entity type defines an ordered basis: $\mathcal{T}^{e} \triangleq (\mathcal{T}^{f}_{n_i})_{i=1}^{N^e}$ over a vector space $\mathcal{X}_{\mathcal{T}^e}$, where the ordering is fixed by the environment, and the basis elements $\mathcal{T}^{f}_{n_i} \in \mathscr{T}_{\mathit{factor}}$ are factor types, each spanning a vector space, $\mathcal{F}(\mathcal{T}^{e})_i$.
The complete vector space of an entity type then is a product space $\mathcal{X}_{\mathcal{T}^e} =\prod_i \mathcal{F}(\mathcal{T}^{e})_i$.
The set of all available entity types is denoted as $\mathscr{T}_{\mathit{entity}}$.

Given these definitions, an \textbf{entity} in \segar{} is an instance of an entity type, i.e., a specific object, and a \textbf{factor} is an instance of a factor type, i.e., a property of a specific object. E.g., if \textsc{Mass}, \textsc{Radius}, \textsc{X}, and \textsc{Y} are factor types and \textsc{Ball} = (\textsc{Mass}, \textsc{Radius}, \textsc{Position}) is an entity type, then $\textsc{Ball}: B_7 = (\textsc{Mass}:M_7, \textsc{Radius}:R_7, \textsc{Position}: (X_7, Y_7)$ is a specific ball, and $M_7$, $R_7$, $X_7$, and $Y_7$ are factors denoting $B_7$'s mass, radius, and position. 
Therefore, each entity of type $\mathcal{T}^{e}$ is represented by a vector of factor values in $\mathcal{X}_{\mathcal{T}^e}$.

To define the complete \textbf{state space}, let $\mathcal{E}$ be a finite multiset of entity types from $\mathscr{T}_{\mathit{entity}}$, e.g., $(\textsc{Ball}, \textsc{Ball}, \textsc{Magnet}, \textsc{Sand}, \textsc{Sand})$, where the element ordering is fixed by the environment.
Then the complete space defined by $\mathcal{E}$ is a product space, $\mathcal{X} = \prod_{\mathcal{T}^e \in \mathcal{E}} \mathcal{X}_{\mathcal{T}^e}$.
The vectors $x \in \mathcal{X}$ have no special structure outside of the ordering of the entities and factors prescribed by the environment, so states can be interpreted as vectors on a product space over all factor spaces over all entities, $\mathcal{X} = \prod_{\mathcal{T}^{e} \in \mathcal{E}} \prod_i F(\mathcal{T}^{e})_i$. 

Entity interactions are governed by \textbf{rules}, $\mathscr{R}$.
Let $\mathcal{B}$ be the set of all multisets over entity types $\mathcal{T}^{e} \in \mathscr{T}_{entity}$.
Note first that $\mathcal{E} \in \mathcal{B}$ from above, so like $\mathcal{E}$, every $b \in B$ has a associated vector space $\mathcal{X}_b$ over the multiset of entity types.
For $b \in \mathcal{B}$, a rule, $r(b): \mathcal{X}_b \mapsto \mathcal{X}_b$ is a function that applies to and modifies a vector of factor values over entities.
Given a set of entities in the arena, applying the rule, $r(b)$, amounts to finding all combinations of entities with types that match $b$, then modifying their factor values as specified by the rule.
There is additional logic to manage when multiple rules apply to the same entity's factor, and this is covered in \Cref{sec:details}.
Note that a factor type's meaning in the physical sense is set by the rules and is not inherent to the factor types themselves, other than their base type (float, boolean, int, vector, etc): \textsc{Mass} could for example have the same physical meaning as \textsc{Charge}, depending on the rule definitions (see \Cref{sec:details} for example rules).

The entity factors are observable through an \textbf{observation function} $z: \mathcal{P}(\mathscr{T}_{factor}) \times \mathcal{X} \mapsto \mathcal{O}$, where $\mathcal{P}(\mathscr{T}_{factor})$ is the power set over factor types.
Values of factors dictating entity behavior may or may not be directly observable, as $z$ is designed to only admit observations of a subset of all factor types.
This indicates whether different values of a factor of a given type are manifested visually, e.g., as object color.
$z$ can be stochastically drawn from a larger set of observation functions.
For example, a user can define $z$ to such that \textsc{Mass} and \textsc{Charge} are observable, but not \textsc{Friction}. 
The observation function $z$ will select values of factors of type \textsc{Mass} and \textsc{Charge}, excluding \textsc{Friction}, from the states $x \in \mathcal{X}$, transforming these values to values on the observation space.

The final components are the \textbf{action space}, $\mathcal{A}$, and the \textbf{reward function}, $r: \mathcal{X} \times \mathcal{X} \times \mathcal{A} \mapsto \mathbb{R}$.
The action space is typically an input at an action function, $a: \mathcal{A} \times \mathcal{X} \mapsto \mathcal{X}$, which directly operates on the complete space over all entities.
An action can be specified as any function on the states and can override any other rules, such as those specified by the transition function.

Finally, we define a \textbf{task} in \segar{}, which is a tuple $\langle \mathscr{T}_{\mathit{factor}}, \mathscr{T}_{\mathit{entity}}, \{\mathscr{E}_{\mathcal{T}^e}\}_{\mathcal{T}^e \in \mathscr{T}_{\mathit{entity}}}, \mathscr{R}, \mathcal{A}, r, z, c_0 \rangle$, where $c_0$ is the initial configuration of all of the task's entities, i.e. an assignment of values to all entities' factors. For instance, if a task involves one ball and one charged ball, the initial configuration could be  $c_0 = \{(\textsc{mass}:M_1 = 3, \textsc{radius}:R_1 = 0.2, \textsc{X}:X_1 = 5.32, \textsc{Y}:Y_1 = 1.04), (\textsc{mass}:CM_1 = 3, \textsc{radius}:CR_1 = 0.2, \textsc{X}:CX_1 = 2.11, \textsc{Y}:CY_1 = 6.97,  \textsc{charge}:CC_1 = -10)\}$. 
\segar{}'S major contribution are tools for imperatively defining \textbf{task distributions}.
Sampling a task entails first constructing $\mathcal{E}$ by generating a number of entities of each type.
Then, the initial configuration, $c_0$ is drawn by sampling an initial value for every factor of every generated entity from a set of given priors.

\section{Additional \segar{} details}
\label{sec:details}

In this section we will provide examples and more details on \segar{}'s components.
We also recommend following the tutorials (Jupyter notebooks and READMEs) in the repository (\href{https://github.com/microsoft/segar/tree/main/segar}{https://github.com/microsoft/segar/tree/main/segar}) to develop a more comprehensive understanding of how \segar{} works.

\subsection{Factor Types}

As mentioned in \Cref{sec:formal}, the core concept in \segar{} is the set of factor types used to build tasks.
Each factor type has an associated \emph{base type} in its definition that dictates what values it can take.
Some examples of the factor types include:
\begin{itemize}
    \item \textsc{Position}: A 2D vector representing the location of an entity within the arena.
    \item \textsc{Velocity}: A 2D vector representing the changes in position, to integrate over time, of an entity within the arena (see the \textsc{move} rule below).
    \item \textsc{Charge}: A float-valued factor that specifies charge of the entity. When the Lorentz law of electromagnetic force~\citep{panofsky2005classical} is used as a rule (see rules below), this will affect the acceleration of all entities with charge.
    \item \textsc{Mass}: A float-valued factor that specifies the mass. For all physical interactions (rules or actions), this value is inversely proportional to how much force needs to be applied to accelerate an entity.
    \item \textsc{Friction}: A float-valued factor that, when friction rules are applied, represents the kinetic friction coefficient of a \textsc{Tile} (see below) used to compute the force due to friction applied to another entity (such as an \textsc{Object}).
    \item \textsc{Alive}: A boolean-valued factor that, can indicate to rules whether any other factors within the same entity should be applied or not (e.g., entities that are not alive should not contribute to the Lorentz force).
\end{itemize}

Note again that the factor types above have \emph{physical meaning} based on the rules, which the researcher can specify and customize to fit their research needs.
New factor types can be introduced through subclassing the generic \textsc{Factor} type or subclassing an existing factor type.
Any subclass of a factor type will be applied to rules of the parent, but not visa-versa.
A tutorial on factors can be found in the repository (\href{https://github.com/microsoft/segar/tree/main/segar/factors}{https://github.com/microsoft/segar/tree/main/segar/factors}).

\subsection{Entity Types}

A set of factor types \emph{defines} an entity type, with exception to the generic \textsc{Entity} type.
Entities are instantiated with dictionaries with factor types as keys.
For example, the dictionary representation for an object might look like:
\begin{lstlisting}[language=Python]
{
    Position: [0.1, -0.1], 
    Mass: 0.4, 
    Velocity: [0.1, 1.0], 
    Shape: Circle,
    # Shapes such as Circle are special classes 
    # within SEGAR that have special collision rules
    ... 
}
\end{lstlisting}
Some examples of built-in entity types include:
\begin{itemize}
    \item \textsc{Entity}: A special abstract container of factors (generic type). Can be thought of as a dictionary with factor types as key entries and the corresponding factor values as the values. \textsc{Entity} is the only entity type without a set of factor types in its definition (can use any factory type as dictionary keys).
    \item \textsc{Thing}: An entity with position, shape, and size factors. These correspond to the class of entities that are ``localizable" in the arena, i.e., through their position, location, and shape.
    \item \textsc{Object}: A thing that has mass, charge, and other object-like factors, but most importantly have velocity. 
    Objects move around in the arena and collide with walls and other objects.
    \item \textsc{Charger}: A object with charge (i.e., will respond to the Lorentz force).
    \item \textsc{Tile}: A thing that does not move nor collide, but can apply force to objects that overlap with them. Objects move over tiles.
    \item \textsc{Sand}: A tile with friction, which when the \textsc{apply\_friction} rule is in play, will de-accelerate objects on top of them.
\end{itemize}

\subsection{Rules}
Rules define the transition function of the environment as well as implement the \emph{physics} of the task. 
\segar{} provides full control over which rules apply to what entities and how (i.e., through the factor types).
For example, the following rule applies friction on one entity from another entity:

\newpage
\begin{lstlisting}[language=Python]
def apply_friction(
        o1: Tuple[Mass, Velocity, Acceleration],
        o2: Tuple[Friction],
        g: Gravity) -> Aggregate[Acceleration]:

    m, v, a = o1
    mu, = o2
    if v.norm() >= 1e-6: 

        v_sign = v.sign()
        v_norm = v.norm()

        f_mag = mu * g
        norm_v = v.abs() / v_norm
        da = -v_sign * f_mag * norm_v / m
        return Aggregate[Acceleration](a, da)
    else:
        return None
\end{lstlisting}

\segar{} rules make use of the Python type hints system (\href{https://docs.python.org/3/library/typing.html}{https://docs.python.org/3/library/typing.html}), which can be used specify the input types in the function signature.
For \segar{}, this is required if a function is to be used as a rule.
In addition, the special \href{https://docs.python.org/3/library/typing.html#typing.Tuple}{\textsc{Tuple}} type is used to specify an abstract entity type, with the items in the square brackets being the factor types of said entity type.
In the above example, the rule applies to two entities: one that contains mass, velocity, and acceleration, and the other that contains friction. 
The simulator, when it applies rules, will search for all possible combinations of pairs of distinct entities with these factors, inclusive to those entities containing other factor types as well.
In other words, if a rule applies to an entity type, $\mathcal{T}^e$ with factor types $\mathscr{T}^e_{factor}$, it will apply to a sub-class of $\mathcal{T}^e$, $\mathcal{T}^{e'}$ with factor types $\mathscr{T}^{e'}_{factor} \supseteq \mathscr{T}^e_{factor}$.

The body of the rule is written in Python, where all operations (e.g., multiplication) are defined by the factors (by default the operations are the same as the base type of the factor type).
The return type in the example above, \textsc{Aggregate}, is a special type that indicates to the simulator what to do with the output of the function (see below), and it's item is the factor type of the output.
Return types can also be \textsc{Tuple}, to indicate the change of additional factors.

Overall, there are four return types for rules in \segar{}:
\begin{itemize}
    \item \textsc{Aggregate}: All rules that apply to the same factor and that have this type should be added, forgetting the previous value.
    \item \textsc{Differential}: The result is scaled by the time-scale and added to the prior value.
    \item \textsc{SetFactor}: This overrides all other rule types and sets the factor to the output value.
    \item \textsc{None}: This rule has no effect.
\end{itemize}
As another example, consider the following rule for the Lorentz force:
\newpage
\begin{lstlisting}[language=Python]
def lorentz_law(
    o1_factors: Tuple[Position, Velocity, Charge, Magnetism],
    o2_factors: Tuple[Position, Velocity, Charge, 
                      Mass, Acceleration],
) -> Aggregate[Acceleration]:

    x1, v1, q1, b1 = o1_factors
    x2, v2, q2, m2, a = o2_factors
    normal_vec = x2 - x1
    unit_norm = normal_vec.unit_vector()

    if m2 == 0:
        # No mass, no acceleration.
        return Aggregate[Acceleration](a, 0.0 * a)

    if q1:
        f_q = q1 * q2 * unit_norm / normal_vec.norm() ** 2
    else:
        f_q = 0.0

    rel_vel = Velocity(v2 - v1)
    if b1 and rel_vel.norm() != 0.0:
        unit_vel = Velocity(rel_vel.unit_vector())
        tang_vel = unit_vel.tangent_vector()
        f_b = (q2 * rel_vel.norm() * b1 * tang_vel / 
               rel_vel.norm() ** 2)
    else:
        f_b = 0.0

    da = (f_b + f_q) / m2
    
    return Aggregate[Acceleration](a, da)
\end{lstlisting}

Suppose there are three objects in the arena, $o_1$, $o_2$, $o_3$, each with position, velocity, charge, and magnetism. 
The simulator will apply the Lorentz force to $o_1$ (resp. $o_2$ and $o_3$), using the charge and magnetism from $o_2$ and again from $o_3$.
The ouputs of these rules are \emph{aggregated} to determine the final acceleration to $o_1$.
This is instantaneously applies to the acceleration of $o_1$.

To contrast, the move rule:

\begin{lstlisting}[language=Python]
def move(
        x: Position, v: Velocity, 
        min_v: MinVelocity
        ) -> Differential[Position]:
    if v.norm() < min_v:
        dx = 0 * v
    else:
        dx = v
    return Differential[Position](x, dx)
\end{lstlisting}

is sensitive to time scale of the environment, and represents discrete approximation of the integral:

\begin{equation*}
    x(t_1) = x(t_0) + \Delta{x}; \, \Delta{x} = \int_{t_0}^{t_1} v(t) \,dt \approx v(t_0) (t_1 - t_0).
\end{equation*}
\textsc{Differential} returns on the same factor will aggregate to the prior value.

For conflict resolution, \textsc{SetFactor} will override all other rule types, and \textsc{Aggregate} will override \textsc{Differential}.
If there are more than one \textsc{SetFactor} applied to the same factor, then additional logic is used to determine which rule to apply based on how specific the rule is (e.g., is it conditioned on other factors of the same entity). If the logic ultimately can not resolve a rule conflict, then this will result in an error.
This apply-logic-or-fail property in \segar{} ensures than any stochasticity in the transition function is not due to how the simulator applies the rules.
Any stochasticity in the transition function \emph{must be implemented by the researcher}, e.g., can only be from the rule functions themselves (e.g., by using a random number generator) or from any randomness from the factors (which the researcher also controls).
Finally, there is a set of special rules, collisions, which are always applied last after position factors have changed.

The set of built-in rules \segar{} provides are found in the \href{https://segar.readthedocs.io/en/latest/autoapi/segar/rules/}{documentation} with a tutorials provided in the repository (see \href{https://github.com/microsoft/segar/tree/main/segar/rules}{https://github.com/microsoft/segar/tree/main/segar/rules} or \href{https://github.com/microsoft/segar/tree/main/segar/sim}{https://github.com/microsoft/segar/tree/main/segar/sim} for examples).
However, like the factor and entity types above, the build-in rules are optional and extensible: the researcher may provide any set of rules they would like to the simulator, as long as they follow the same type hints system as the built-in ones.

\subsection{Observation function}

\begin{figure}
    \centering
    \includegraphics[width=.8\textwidth]{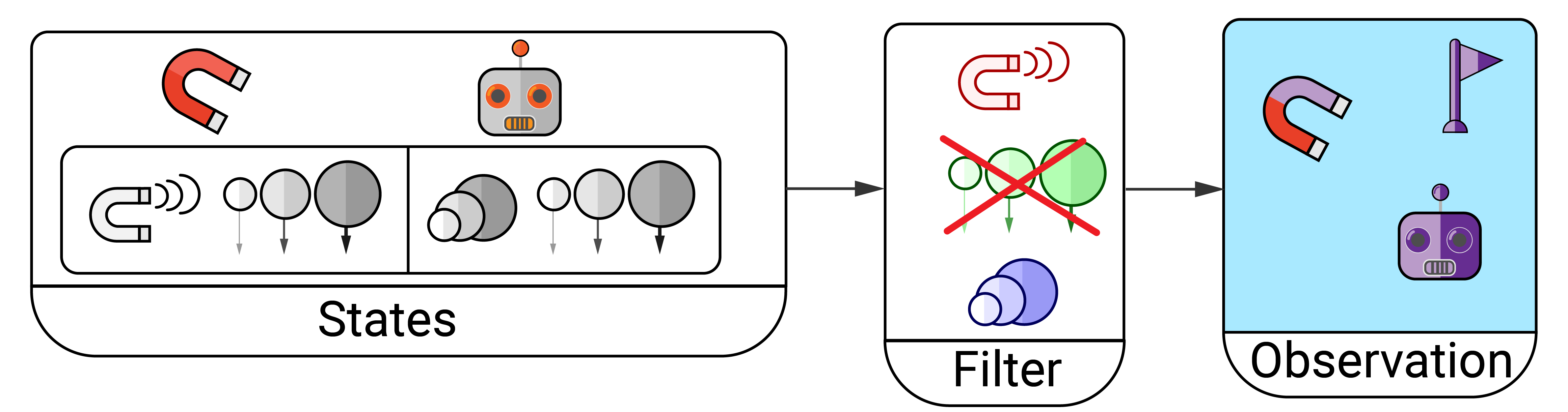}
    \label{fig:observations}
    \caption{
    The observation function in \segar{} can be customized by the researcher as any function of the input states. Addition functionality for filtering factors given a type of observation function (such as state- or pixel-based) gives additional control over whether the problem is a MDP or a POMDP with respect to specific factors.}
\end{figure}

The observation function is a fully-customizable component of \segar{} that transforms the states into observations seen by the agent.
Observation functions can be state-based or anything else derived from states (such as pixel-based observations), can be multimodal, and can provide partial observability.

\begin{figure}
\centering
\begin{subfigure}[b]{0.38\textwidth}
  \centering
  \includegraphics[width=.49\textwidth]{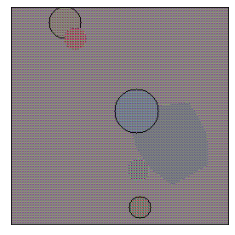}
  \includegraphics[width=.49\textwidth]{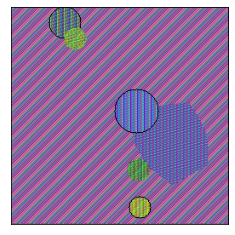}
  \caption{}
  \label{fig:vis_features}
\end{subfigure}
\hfill
\begin{subfigure}[b]{0.59\textwidth}
  \centering
  \includegraphics[width=.32\textwidth]{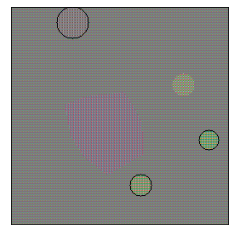}
  \includegraphics[width=.32\textwidth]{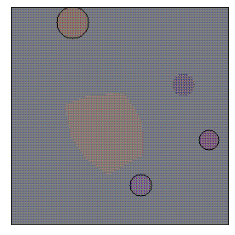}
  \includegraphics[width=.32\textwidth]{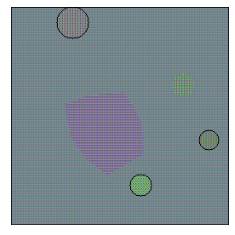}
  \caption{}
  \label{fig:vis_features2}
\end{subfigure}
\caption{
(a) Two example visual renderings of the state state from \segar{}.
Colors and patterns correspond to different factors and their values, with one pattern corresponding to the background.
Left) Visual features from a trained auto-encoder where the input is drawn from the product of marginal uniform distributions over all factors. The agent should be nearly guaranteed to be able to infer all factor values from visual features.
Right) Visual features derived from clustering the weights from the first layer of an Inception network.\\
(b) Provided variants of linear auto-encoder visual features allows the user to vary the observation space, challenging the agent's ability to adapt to environments with the same physics, yet different visual features. These are generated from different auto-encoders trained with different seeds. Similar variants are provided for the the Inception renderer.
}

\end{figure}

For the pixel-based observations, \segar{} provides built-in \emph{renderers} that generate visual features from the underlying factors.
Two examples are provided in Figure~\ref{fig:vis_features}.
The first built-in renderer is trained from scratch using an autoencoder, with an encoder that takes samples from the marginal distribution over factor types as input and outputs visual features in RGB pixel space.
The autoencoder is given a reconstruction network and the full model is trained to reconstruct samples from the product of marginals.
The second built-in renderer is constructed from the first layer weights from an Inception network trained on Imagenet~\citep{szegedy2017inception} along with a k-means to cluster these weights, randomly associating each cluster with a factor type.
We provide sets of both of these renderers, generated with different initial seeds, allowing for measurable variation across observation spaces (see Figure~\ref{fig:vis_features2} for examples of the auto-encoder renderer given the same state).
A tutorial on the renderers as well as code that can reproduce the visual features of these renderers can be found in the repository (\href{https://github.com/microsoft/segar/tree/main/segar/rendering}{https://github.com/microsoft/segar/tree/main/segar/rendering}).

\subsection{Task initialization and sampling distributions}
\begin{figure}
    \centering
    \begin{subfigure}[b]{0.45\textwidth}
    \centering
    \includegraphics[width=\textwidth]{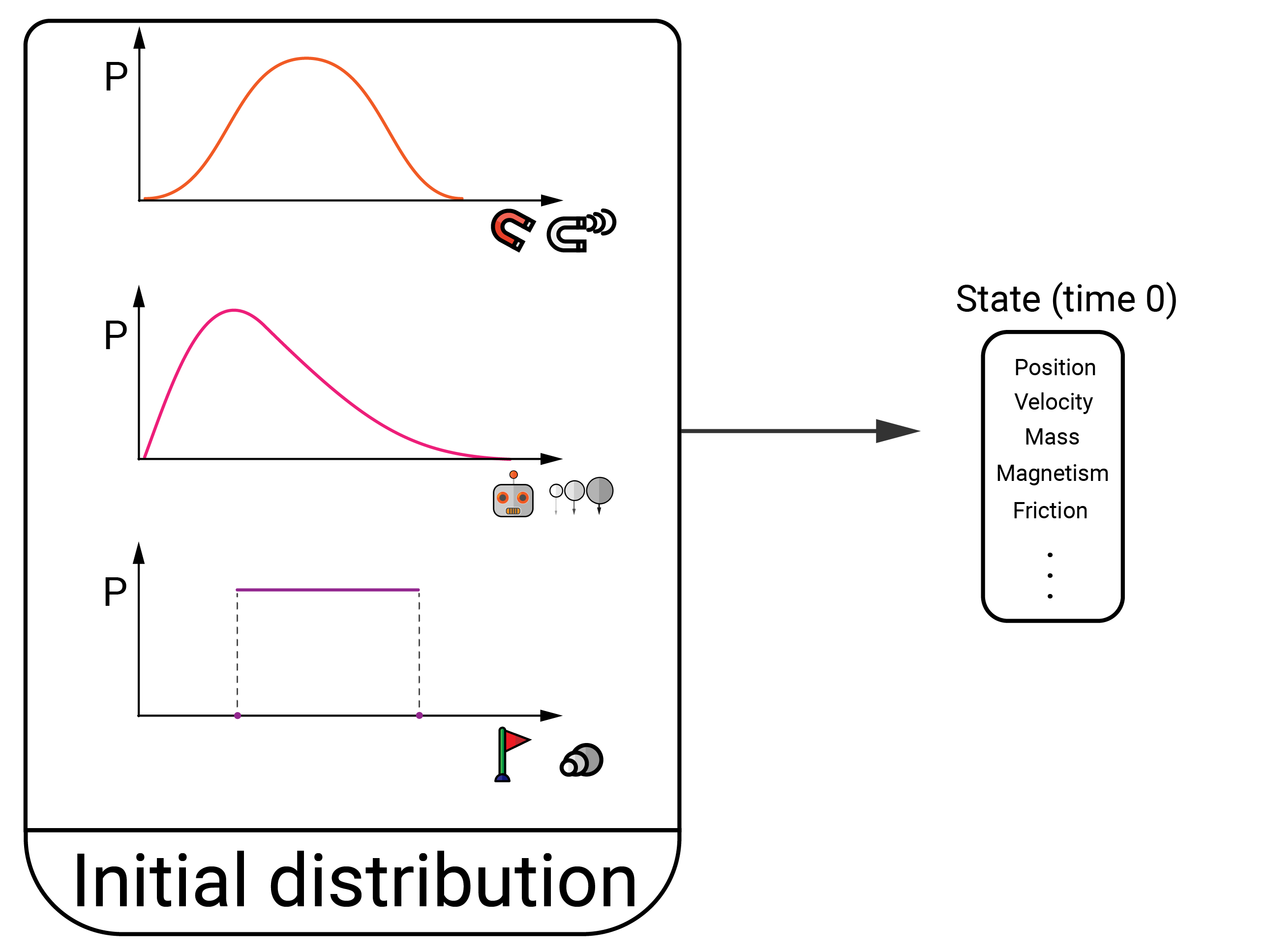}
    \caption{}
    \label{fig:distributinos}
    \end{subfigure}
    \hfill
    \begin{subfigure}[b]{0.45\textwidth}
    \centering
    \includegraphics[width=\textwidth]{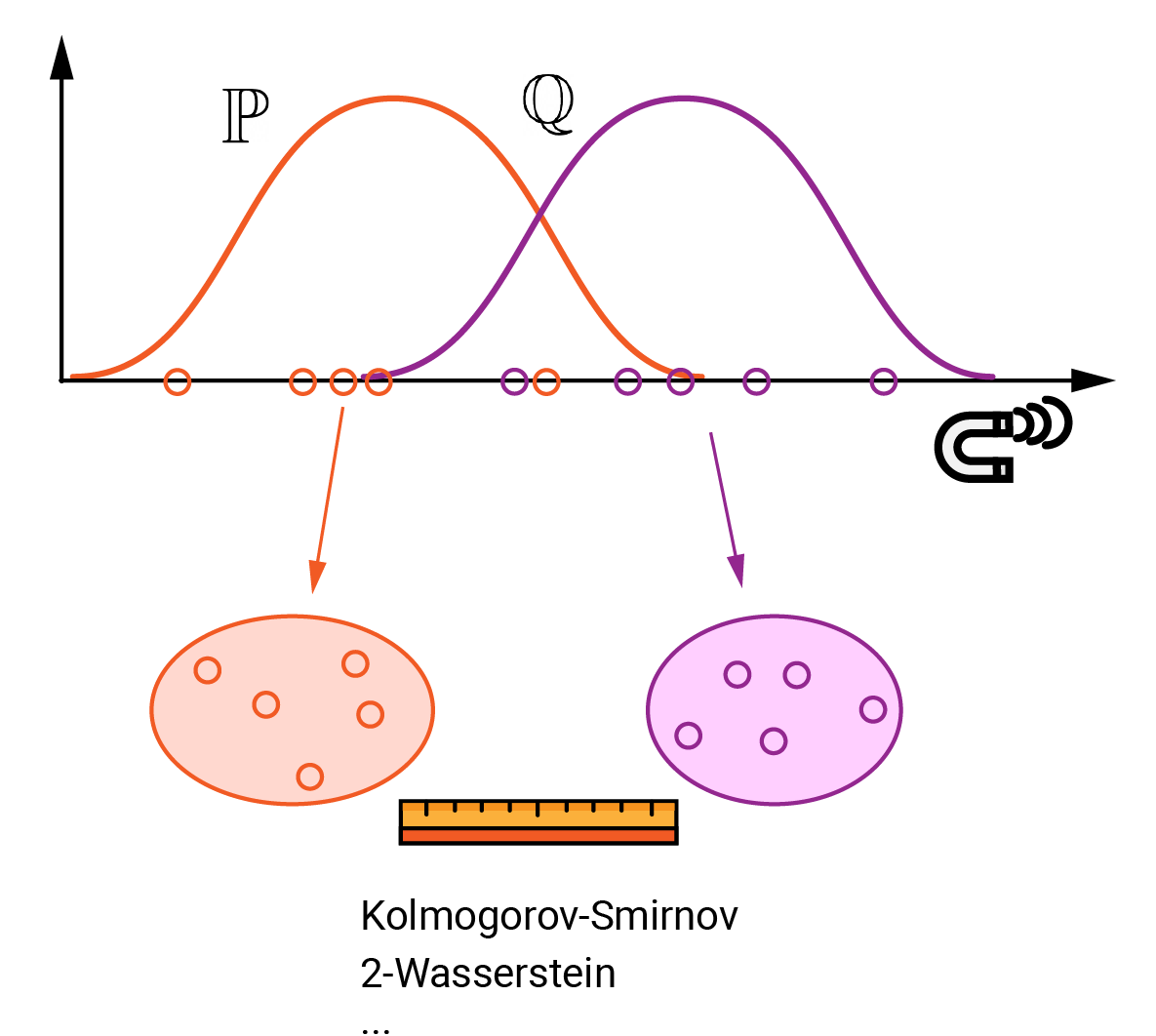}
    \caption{}
    \label{fig:measures}
    \end{subfigure}
    \caption{
    (a) \segar{} allows for specification of the distribution from which tasks are drawn, in the above case in terms of initializations. \\
    (b) As the underlying factors are exposed and their underlying distributions known, there are many ways to measure the distance between sets of tasks as well as between sets of tasks and task distributions.
    This includes Kolmogorov-Smirnov (sample to distribution) and Wasserstein-2 (sample to sample) distances.}
\end{figure}

Distributional control and transparency represent core functionalities of \segar{} that allows for controlled experimentation in generalization for \lint{}.
This is done through the distributions of entity types (their numbers) and the values of their respective factors at initialization, and an illustration is provided in \Cref{fig:distributinos}.
Some examples of how distributions can effect initialization are provided in the MDP (\href{https://github.com/microsoft/segar/tree/main/segar/mdps}{https://github.com/microsoft/segar/tree/main/segar/mdps}), task (\href{https://github.com/microsoft/segar/tree/main/segar/tasks}{https://github.com/microsoft/segar/tree/main/segar/tasks}), and factor (\href{https://github.com/microsoft/segar/tree/main/segar/factors}{https://github.com/microsoft/segar/tree/main/segar/factors}) tutorials.

For example, the default initialization configuration for \textsc{PuttPutt} is a dictionary with the following entries:
\newpage
\begin{lstlisting}[language=Python]
puttputt_default_config = {
    "numbers": [
        (ThingFactory([Charger, Magnet, Bumper, Damper, Ball]), 
         DiscreteRangeNoise(1, 2)),
        (
            ThingFactory({SandTile: 2 / 5.0, MagmaTile: 1 / 5.0,
                          Hole: 1 / 5.0, FireTile: 1 / 5.0}),
            DiscreteRangeNoise(1, 2),
        ),
        (GoalTile, 1),
        (GolfBall, 1),
    ],
    "priors": [
        Prior(Position, RandomEdgeLocation(), 
              entity_type=Object),
        Prior(Position, RandomMiddleLocation(),
              entity_type=Tile),
        Prior(Position, RandomBottomLocation(),
              entity_type=GolfBall),
        Prior(Position, RandomTopLocation(),
              entity_type=GoalTile),
        Prior(Shape, RandomConvexHull(0.3), entity_type=Tile),
        Prior(Shape, Circle(0.3), entity_type=GoalTile),
        Prior(Shape, Circle(0.4), entity_type=Hole),
        Prior(Size, GaussianNoise(0.2, 0.01, clip=(0.1, 0.3)), 
              entity_type=Object),
        Prior(Size, GaussianNoise(1.0, 0.01, clip=(0.5, 1.1)), 
              entity_type=Tile),
        Prior(Mass, 1.0),
        Prior(Mobile, False),
        Prior(Mobile, True, entity_type=Ball),
        Prior(Mobile, True, entity_type=GolfBall),
        Prior(
            Charge, GaussianMixtureNoise(means=[-1.0, 1.0], 
            stds=[0.1, 0.1]), entity_type=Charger
        ),
        Prior(
            Magnetism, GaussianMixtureNoise(means=[-1.0, 1.0], 
            stds=[0.1, 0.1]), entity_type=Magnet
        ),
        Prior(Friction, UniformNoise(0.2, 1.0), 
              entity_type=SandTile),
    ],
}
\end{lstlisting}

This configuration contains information about 1) the number of entity types, specified by a tuple of the type and number or a type factory and a distribution, and 2) the values of the factors, specified by a \textsc{Prior} or value.
The noise types, such as \textsc{GaussianNoise}, \textsc{DiscreteRangeNoise}, and \textsc{UniformNoise}, are special factor types that can be sampled from.
At initialization, the \textsc{Prior} is matched to its corresponding factor by conditioning variables and either sets the value if it is fixed or samples in the case the value is a noise factor.
For instance:
\begin{lstlisting}[language=Python]
    Prior(Size, GaussianNoise(0.2, 0.01, clip=(0.1, 0.3)), 
          entity_type=Object)
\end{lstlisting}
samples the value of all size factors, conditioned on that the size belongs to an entity with type \textsc{Object}.
This schema provides fine-grained programmatic control over initialization of tasks, which in turn directly links generalization experiment design to known parametric distributions.

Conflict resolution is done by specificity of the conditions of the prior.
For example, with the initialization:

\newpage
\begin{lstlisting}[language=Python]
    Prior(Size, GaussianNoise(0.2, 0.01, clip=(0.1, 0.3)), 
          entity_type=Object),
    Prior(Size, GaussianNoise(0.3, 0.02, clip=(0.1, 0.5)), 
          entity_type=Charger),
\end{lstlisting}
the prior with condition \lstinline{entity_type=Charger} is more specific, as \textsc{Charger} is a subclass of \textsc{Object}, so in the case of the initialization of a \textsc{Charger}, which is also an \textsc{Object}, only the second rule would apply.
If no resolution is possible (e.g., in cases of priors with identical conditions), this will result in an error.

\subsection{Task measures on distributions and samples}
As task initialization are drawn from and specified by distributions, these distributions can be used to measure properties of an initialization (such as entropy), distances between different initializations, and how representative samples are from a given distribution (an simple illustration is provided in \Cref{fig:measures}).
The distributions are instantiated through the noise factors, and the noise factors provide a number of statistical methods to aid in measuring, such as CDFs, PDFs, entropy, etc.

To assess how close the training distribution is to the test distribution, one typically needs to compute the divergence between both distributions. This area has been extensively studied in a panoply of previous works, which propose multiple divergences, analyzing their population and sample-based estimators. In the end of the day, the choice of the divergence should take into account two properties: expressivity of the metric for any general family of distributions, and the sample size required to estimate it up to some error. 

Guided by this principle, we chose the Wasserstein-2 metric and the Kolmogorov-Smirnov statistic for our experiments.

\paragraph{Wasserstein-2} We use unbalanced optimal transport~\citep{bonneel2011displacement, bonneel2015sliced} to compute the distance between two sets of tasks (e.g., train vs test).
In order to do this, we must take a hierarchical approach to solving the optimal transport problem, as any transport across tasks must match across entities and their respective factors.
The pseudocode for computing this distance across sets of tasks is given in \Cref{alg:W2}, and we implement $W_2(.,.)$ and $\text{EMD}(.)$ using the Python Optimal Transport library~\citep[POT, \href{https://pythonot.github.io}{https://pythonot.github.io}][]{flamary2021pot}.

\begin{algorithm}[H]
 \SetAlgoLined
 \SetKwInOut{Inputs}{Inputs}
 \SetKwInOut{Hyperparameters}{Hyperparameters}
 \SetKwInOut{Given}{Given}
 \Inputs{
 Two sets of tasks, $T_1$ and $T_2$.
 Each task $t \in T$ has a set of entities at initialization $t.\text{entities}$.
 Each set of entities $E \subseteq t.\text{entities}$ also has set of initial vector states $E.\text{states} = \{e.\text{state}\}_{e \in E}$.
}
 \Hyperparameters{$m$ -- entity type mismatch distance}
 \Given{
 $W_2(.,.)$ -- a method for estimating Wasserstein-2 across two sets of vectors of identical length.
 $\text{EMD}(.)$ -- A solver for the unbalanced Earth Movers Distance (EMD) problem given a cost matrix.
 }
 $C \gets$ zero array with $len(T_1)$ rows and $len(T_2)$ columns \tcp{cost matrix}
 $E\text{types} \gets []$ \tcp{list of all entity types}
\For{$t \in T_1 \cup T_2$}
    {
    \For{$e \in t.\text{entities}$}
        {
        \If{$\text{type}(e) \notin E\text{types}$}
            {$E\text{types}.\text{append}(\text{type}(e))$}
        }
    } 
 \For{$t_i \in T_1$}{
    \For{$t_j \in T_2$}
        {
        \tcp{extract entities at initialization}
        $E_i \gets t_i.\text{entities}$\\
        $E_j \gets t_j.\text{entities}$\\
        \For{$d \in \text{type}(e)$}
            {
            \tcp{collect sets of entities with the same type}
            $F_i \gets \{e \in E_i | \text{type}(e) = d\}$ \\
            $F_j \gets \{e \in E_j | \text{type}(e) = d\}$ \\
            \If{$|F_i| = 0 \land |F_j| = 0$}
                {
                pass \tcp{ignore tasks with none of this type}
                }
            \ElseIf{$|F_i| = 0 \lor |F_j| = 0$}
                {
                $C[i][j] \gets m$ \tcp{Tasks have entity type mismatch}
                } 
            \Else
                {
                \tcp{adding squared distances across entity types}
                $C[i][j] \gets C[i][j] + W_2(F_i.\text{states}, F_j.\text{states})^2$ 
                }
            }
        }
    }
    \Return $\sqrt{\text{EMD}(C})$
    
  \caption{Wasserstein-2 approximation for task initializations}
  \label{alg:W2}
 
 \end{algorithm}

\paragraph{Kolmogorov-Smirnov} The Kolmogorov-Smirnov (KS) statistic for ground truth cumulative distribution $F=\mathbb{P}[X\leq x]$ and $i.i.d.$ samples $X_1,..,X_n\sim F$ is defined as
\begin{equation}
    \text{KS}_F=\sup_{x}\bigg|\frac{1}{n}\sum_{i=1}^n\mathbb{I}[X_i\leq x]-F(x)\bigg|
\end{equation},
which quantifies how far the empirical cumulative distribution lies from the true CDF; the gap shrinks at a rate exponential in $n$. This statistic can be easily adapted to handle population quantities, so as to measure the distance between CDFs $F$ and $G$:
\begin{equation}
    \text{KS}_{F,G}=\sup_{x}\bigg|G(x)-F(x)\bigg|\;.
\end{equation}
The quantity $\text{KS}_{F,G}$ is \emph{not} a random variable, but is an exact measure of the largest discrepancy of both CDFs. The main advantage of \segar{} over other benchmarks is that it lets us compute $\text{KS}_{F,G}$, as all factor distributions are known exactly. That is, in practice, we do not necessarily know the cross-dependence structure of factors (although it can be estimated). We therefore rewrite the probability of a task defined by a realization of environment factors $x_1,..,x_k$ as the product of the marginal distributions:
\begin{equation}
    F(x_1,..,x_k)\approx \prod_{j=1}^k\mathbb{P}[X_j=x_j],
\end{equation}
which significantly simplifies the computation of $\text{KS}_{F,G}$.
Pseudocode for our KS statistic computation is provided in \Cref{alg:KS}.

\begin{algorithm}[H]
 \SetAlgoLined
 \SetKwInOut{Inputs}{Inputs}
 \SetKwInOut{Hyperparameters}{Hyperparameters}
 \SetKwInOut{Given}{Given}
 \Inputs{
 Two sets of tasks, $T_1$ and $T_2$.
 Each task $t \in T$ has a set of entities at initialization $t.\text{entities}$.
 Each entity type, $\text{type}(e)$ has a associated vector space $\text{support}$.
}
 $E\text{types} \gets []$ \tcp{list of all entity types}
\For{$t \in T_1 \cup T_2$}
    {
    \For{$e \in t.\text{entities}$}
        {
        \If{$\text{type}(e) \notin E\text{types}$}
            {$E\text{types}.\text{append}(\text{type}(e))$}
        }
    } 
\For{$e \in E\text{types}$}
    {
    $F_{1,type(e)}(x)\gets\frac{1}{|T_1|}\sum_{i=1}^{|T_1|}\mathbb{I}[e\leq x], \; x\in \text{support}(e)$ \tcp{Estimate eCDF}
}
\For{$e \in Etypes$}
    {
    $F_{2,type(e)}(x)\gets\frac{1}{|T_2|}\sum_{i=1}^{|T_2|}\mathbb{I}[e\leq x], \; x\in \text{support}(e)$ \tcp{Estimate eCDF}
} 
\For{$e \in E\text{types}$}{
 \tcp{Estimate mean-field eCDFs}
 $F_1(x)\gets \prod_{type(e) \in E\text{types}}F_{1,type(e)}(x),\; x \in support(E\text{types})$ 
}

\Return $\sup_{x\in support(E\text{types})}|F_1(x)-F_2(x)|$
    
  \caption{Two-sample Kolmogorov-Smirnov statistic for task initializations}
  \label{alg:KS}
 
 \end{algorithm}

\section{Additional experiments}

\begin{figure}[ht!]
    \centering
    \includegraphics[width=.5\textwidth]{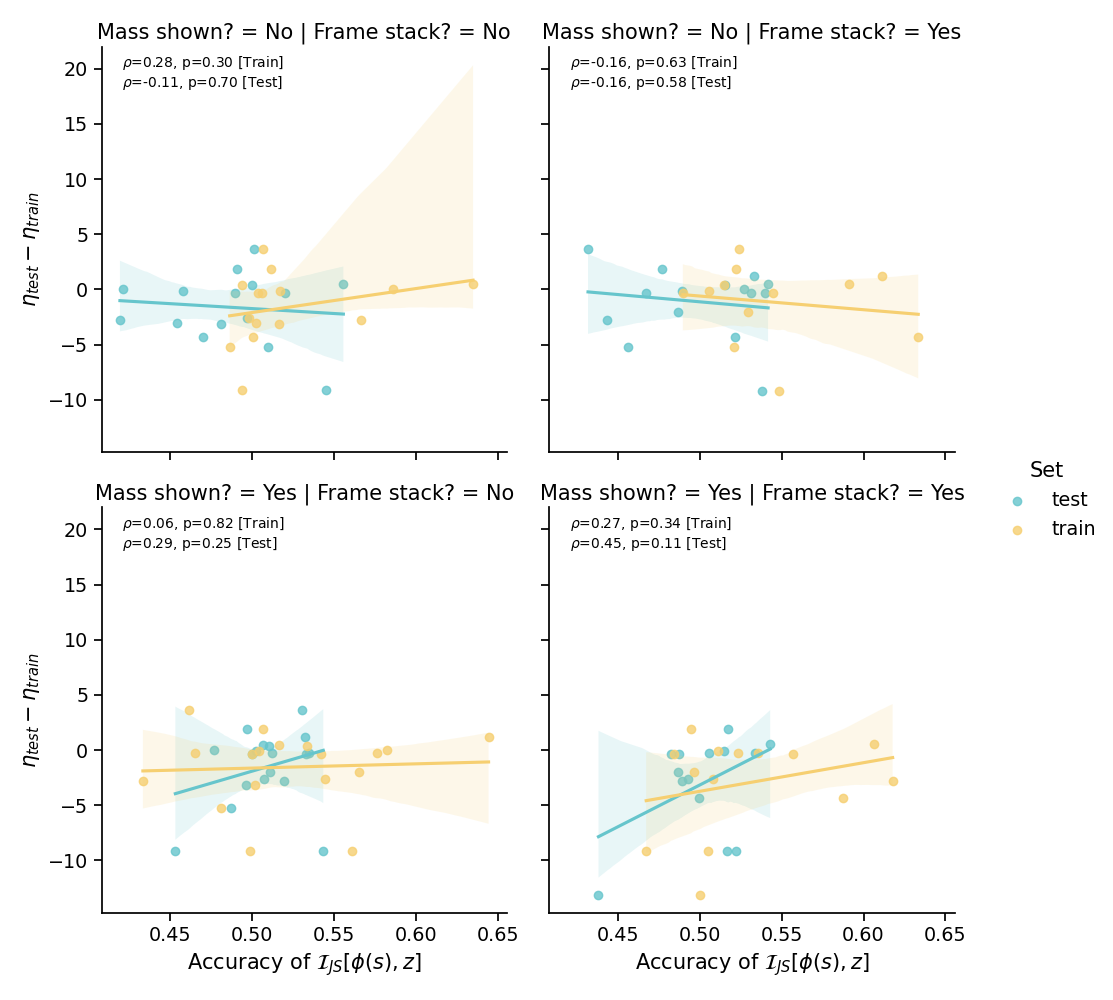}
    \caption{
    Generalization performance gap of a PPO agent as a function of the mutual information between learned state representations and environment factors, when including/ hiding the mass values, as well as adding/removing frame stacking. Each dot represents an experiment on a different number of training levels.}
    \label{fig:returns_vs_mine_fs_mass2}
\end{figure}

\paragraph{What happens when we omit critical factor information (e.g. mass) from observations?} So far, the tasks-specific factors have been fully observable through the feature mapping in the image embeddings of each entity. However, this scenario is highly unrealistic - it is rare that we can directly observe mass, friction, gravity, elasticity of real-world objects. More often than not, a handful of the factors can be sensed directly, while the rest is hidden and has to be extracted from observations. We simulated such a setup in the following experiment, where the mass factor for all entities was removed from the feature maps, making the task specification partially observable. One way to remedy to this issue is, at least in theory, to let the agent observe a sequence of observations via frame stacking. This lets the agent keep track of a "belief state" which can allow the agent to figure out the missing latent factor values based on the dynamics summarized in those observations. 

Figure~\ref{fig:returns_vs_mine_fs_mass2} show the results of this experiment. In particular, note how knowing the mass flips the performance gap trends (within the same task distribution), and helps achieve lower gap value for higher mutual information accuracies.

\begin{figure}[h!]
    \centering
    \begin{subfigure}[b]{\textwidth}
    \centering
    \includegraphics[width=\textwidth]{figures/05_PPO_Returns.png}
    \caption{}
    \label{fig:returns_vs_frames_train2}
    \end{subfigure}\\
    \begin{subfigure}[b]{\textwidth}
    \centering
    \includegraphics[width=\linewidth]{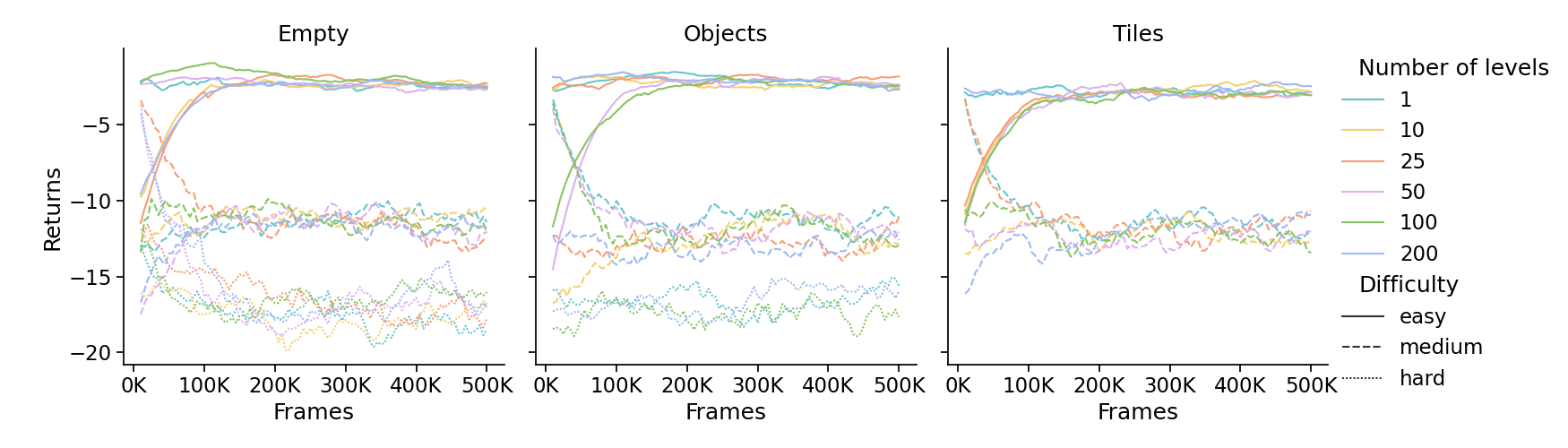}
    \caption{}
    \label{fig:sac_return_curves}
    \end{subfigure}
    \caption{(a) Performance of a PPO agent as a function of training frames. Each curves represents a different task, difficulty and number of levels configuration.\\
    (b) Performance of a SAC agent as a function of training frames. Each curves represents a different task, difficulty and number of levels configuration.}
\end{figure}

\paragraph{What is the effect of training on off-policy data on the performance?} While our main learning algorithm is on-policy (PPO), we investigate how learning from off-policy data affects the agent's performance on \segar{}. To do so, we ran Soft Actor-Critic~\citep{haarnoja2018soft} in a setup identical to the PPO experiment (but using 1 parallel environment instead of 64). Figure~\ref{fig:sac_return_curves} shows that SAC quickly solves the easy tasks as opposed to PPO, but struggles on more challenging configurations.

\section{Additional interesting questions}
\paragraph{Relationship between visual variation and performance}
The generative models for rendering visual features of pixel-based observation spaces are simple with easy-to-access parameters.
That said, we should be able to gauge how different one generator is from another through those parameters, as well as how different sets are.
This control induces a number of experiments similar to those in the previous section, except involving the observation space.

\paragraph{Additional measurements on distributions and sets of tasks}
Among these include measuring properties of sets of tasks, such as entropy or variance, and comparing this measure to performance on the generalization objective.
It would also be interesting to study how specific diversity measures on train and test tasks would reveal important information on the hardness of the generalization objective those task represent.

There are a number of other applications that would demonstrate \segar{}'s potential impact to the \lint{} community as a whole.
In addition to this, \segar{} could be used to develop video benchmarking data as well, potentially making it useful for video comprehension research.
If the reader wishes to discuss their own generalization objectives or related analyses, please feel free to reach out via the GitHub issues.

\section{Additional details}
\subsection{Experimental details}

\label{sec:experiment_details}
\begin{table}[ht]
    \centering
    \begin{tabular}{c|l|c}
        Name & Description & Value \\
        \midrule 
       $n_\text{MDPs}$ & Number of training {POMDPs} & Variable\\
       $\gamma$ & Discount factor & 0.999\\
       Number of environments & Number of environments & 64\\
       Framestack & Number of consecutive frames to stack & 1\\
       Resolution & Observation size (pixels) & 64\\
       Max grad norm & Maximal gradient norm & 10\\
       n-steps & Number of Monte-Carlo steps for GAE estimation & 30\\
       $n_\text{minibatch}$ & Number of PPO batches & 4\\
       $\alpha$ & PPO learning rate & $10^{-4}$\\
       $n_\text{epochs}$ & Number of PPO epochs & 1\\
       $\varepsilon_\text{clip}$ & Clipping range & 0.2\\
       $\lambda$ & GAE lambda & 0.95\\
       $\beta$ & Entropy coefficient & $3\times 10^{-4}$\\
       $c_\text{critic}$ & Critic loss weights & 0.1\\
    \end{tabular}
    \caption{Experiments' parameters}
    \label{tab:appendix_experiment_params}
\end{table}

\subsection{Compute details}
\label{sec:compute_details}

For all experiments, we have used 40 P40 GPUs available on the Azure cloud service.

\subsection{Limitations of \segar{}}
\label{sec:limitations}
While \segar{} provides a first, principled attempt at characterizing the task distribution using a fixed-dimensional set of parameters, the approach is a double-edged sword. The latent distribution over tasks is obtained via the mean-field approximation of all factor distributions which corresponds to the situation where all factors are completely independent of each other. However, this is not always the case, as, for example, "acceleration" and "mass" share a strong causal relationship.

\subsection{Societal impact}
\label{sec:impact}
\segar{} is a generic physics simulator environment that doesn't have any direct connection to real-world systems, therefore no direct negative or positive societal impact is anticipated.
\end{document}